\documentclass[lettersize,journal]{IEEEtran}
\usepackage{amsmath,amsfonts}
\usepackage{algorithmic}
\usepackage{algorithm}
\usepackage{array}
\usepackage[caption=false,font=normalsize,labelfont=sf,textfont=sf]{subfig}
\usepackage{textcomp}
\usepackage{stfloats}
\usepackage{url}
\usepackage{verbatim}
\usepackage{graphicx}
\usepackage{cite}
\usepackage{booktabs}
\usepackage[utf8]{inputenc}
\usepackage{xcolor}
\usepackage{bbding}
\usepackage{pifont}
\usepackage{wasysym}
\usepackage{utfsym}
\usepackage{fontawesome}
\usepackage{multirow}
\usepackage{colortbl}
\usepackage[colorlinks, linkcolor=black, anchorcolor=black, citecolor=black]{hyperref}
\usepackage{soul, color}
\usepackage{bm}

\definecolor{mycolor}{RGB}{0,0,0} 

\definecolor{recolor}{RGB}{0,0,0} 

\begin{document}

\title{Enhancing Environmental Robustness in Few-shot Learning via Conditional Representation Learning}

\author{Qianyu Guo, Jingrong Wu, Tianxing Wu, Haofen Wang, Weifeng Ge, Wenqiang Zhang
\thanks{This work was supported by National Natural Science Foundation of China (No.62072112), the National Natural Science Foundation of China (No.62106051), the National Key R\&D Program of China (No.2022YFC3601405), and Shanghai Municipal Science and Technology Project (Nos. 22N31900400 and 23JC1403400).
\textit{(Corresponding Authors: Weifeng Ge, and Wenqiang Zhang)}}
\thanks{Qianyu Guo and Weifeng Ge are with the School of Computer Science, Fudan University. Qianyu Guo is also with the Shanghai Institute of Virology, Shanghai Jiao Tong University School of Medicine. (e-mail: qyguo20@fudan.edu.cn; wfge@fudan.edu.cn).}
\thanks{Jingrong Wu and Tianxing Wu are with the School of Computer Science and Engineering, Southeast University, Nanjing, 210096, China. (e-mail: 220225818@seu.edu.cn; tianxingwu@seu.edu.cn).}
\thanks{Haofen Wang is with the College of Design and Innovation, Tongji University (carter.whfcarter@gmail.com).}
\thanks{Wenqiang Zhang is with Engineering Research Center of AI \& Robotics, Ministry of Education, Academy for Engineering \& Technology, Fudan University, Shanghai, 20043, China. (e-mail: wqzhang@fudan.edu.cn).}
}

\maketitle

\begin{abstract}
Few-shot learning (FSL) has recently been extensively utilized to overcome the scarcity of training data in domain-specific visual recognition.
In real-world scenarios, environmental factors such as complex backgrounds, varying lighting conditions, long-distance shooting, and moving targets often cause test images to exhibit numerous incomplete targets or noise disruptions.
However, current research on evaluation datasets and methodologies has largely ignored the concept of ``environmental robustness'', which refers to maintaining consistent performance in complex and diverse physical environments. This neglect has led to a notable decline in the performance of FSL models during practical testing compared to their training performance.
To bridge this gap, we introduce a new real-world multi-domain few-shot learning (RD-FSL) benchmark, which includes four domains and six evaluation datasets. The test images in this benchmark feature various challenging elements, such as camouflaged objects, small targets, and blurriness.
Our evaluation experiments reveal that existing methods struggle to utilize training images effectively to generate accurate feature representations for challenging test images.
To address this problem, we propose a novel conditional representation learning network (CRLNet) that integrates the interactions between training and testing images as conditional information in their respective representation processes. The main goal is to reduce intra-class variance or enhance inter-class variance at the feature representation level.
Finally, comparative experiments reveal that CRLNet surpasses the current state-of-the-art methods, achieving performance improvements ranging from 6.83\% to 16.98\% across diverse settings and backbones. The source code and dataset are available at https://github.com/guoqianyu-alberta/Conditional-Representation-Learning.
\end{abstract}


\section{Introduction}
\label{sec:intro}

\begin{figure}[t]
	\centering
	\includegraphics[width=1.0\linewidth]{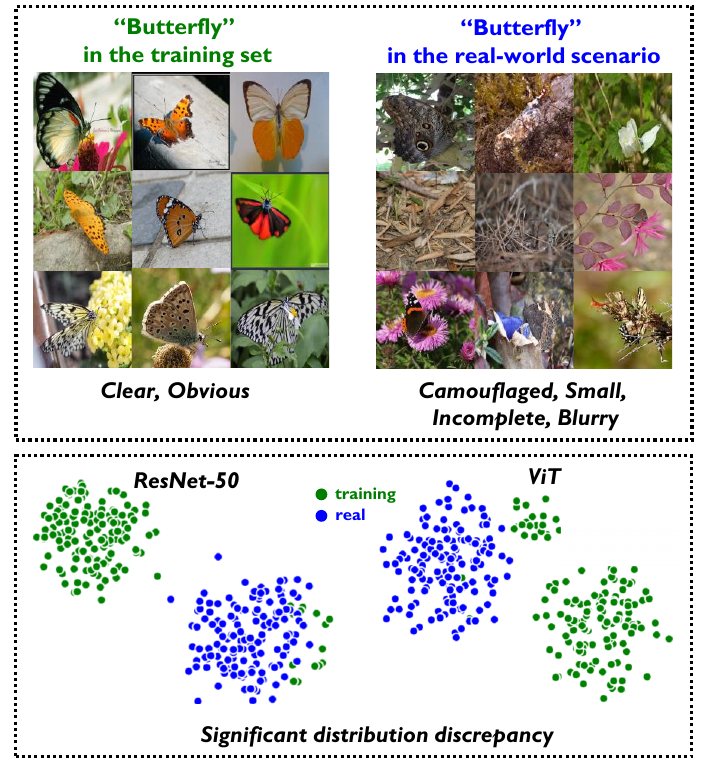}
	\caption{
 Motivation for enhancing ``environmental robustness'' in few-shot learning. Even within the same category, like ``butterflies'', real-world data presents numerous complexities, including camouflaged targets, incomplete targets, and image blurriness, as compared to training data. When features are extracted from both training and real-world data using classical feature extractors like ResNet-50~\cite{HeZRS16} and ViT~\cite{abs-2010-11929}, there is a significant discrepancy in feature distributions for the same category.}
	\label{Fig: motivation}
\end{figure}

\begin{figure}[t]
	\centering
	\includegraphics[width=1.0\linewidth]{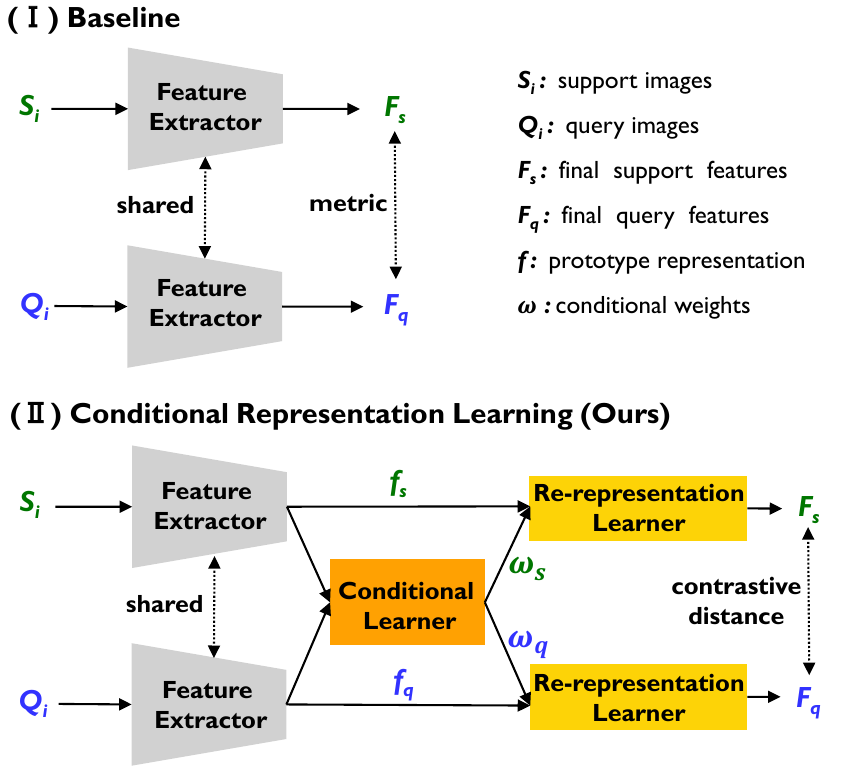}
	\caption{Comparison between the framework of the (\uppercase\expandafter{\romannumeral1}) baseline and (\uppercase\expandafter{\romannumeral2}) the proposed conditional representation learning.
In the proposed conditional representation learning framework, the conditional learner and re-representation learner further optimize the prototype features output by the feature extractor. This enhancement improves the expression of information related to class-discriminative features.}
	\label{Fig: comparison}
\end{figure}

Pre-trained models have surpassed 90\% accuracy in general image classification on ImageNet dataset~\cite{Zhai0HB22, Chen0CPPSGGMB0P23}, yet many categories in downstream visual recognition tasks face challenges due to insufficient annotated training data.
Based on pre-trained models, few-shot learning (FSL)~\cite{VinyalsBLKW16, SnellSZ17} has been proposed to rapidly acquire new meta-knowledge from a limited set of known ``support'' images and generalize to unfamiliar ``query'' images.

Vinyals et al.~\cite{VinyalsBLKW16} initially proposed the MatchingNet, which employs the Siamese Network to separately extract features from support and query images, followed by matching computations on these features. This approach has inspired a variety of subsequent works~\cite{FinnAL17, SnellSZ17, ChenLKWH19, ZhuangTYMZJX18, XieLLWL22, ZhangCLS23, Rizve0KS21, GuoGWFYZG23, FuXFJ23, ZhangGLZFxB24}.
Early efforts in FSL concentrated on training and improving performance on general benchmarks like miniImageNet~\cite{VinyalsBLKW16} and CIFAR~\cite{HeLZZGYZ22}. The latest state-of-the-art (SOTA) FSL methods have achieved approximately 90\% performance on miniImageNet, even in scenarios with only one support image~\cite{Hu0SKH22, HeLZZGYZ22}. 
Recently, several cross-domain benchmarks~\cite{GuoCKCSSRF20, TangWH20} have been introduced to evaluate the generalization capabilities of FSL methods in specific domains, such as medicine and agriculture.
The latest work~\cite{FuXFJ23} achieved over 70\% accuracy on multiple cross-domain datasets, even in the challenging setting with only one support image.

However, in real-world applications, the performance of FSL models often significantly lags behind their performance during the training phase. The acquisition of query images is heavily influenced by various complex environmental factors, such as backgrounds resembling the targets, rapid target movements, and unstable or distant shooting conditions. These factors result in a notable presence of camouflaged, small, or incomplete targets, along with noise and blurriness, as depicted in Figure~\ref{Fig: motivation}.

Popular pre-trained models like ResNet-50~\cite{HeZRS16} and ViT~\cite{abs-2010-11929} struggle to effectively cluster the features of support and query images. Consequently, FSL methods relying on these pre-trained models tend to produce a significant number of classification errors that do not meet practical needs.
Unfortunately, current studies on assessment datasets and methods have overlooked the notion of ``environmental robustness'', which pertains to consistent performance in diverse and demanding physical environments. This oversight has impeded the practical implementation of FSL in the field.

In response to the mentioned challenge, this paper first introduces a novel real-world multi-domain few-shot visual recognition (RD-FSL) benchmark consisting of six datasets spanning domains such as biology, the mining industry, archaeology, and agriculture. Unlike existing evaluation datasets, all images in this benchmark are manually annotated with ``support'' or ``query'' labels. Images labeled as ``query'' must exhibit at least one challenging factor, such as being camouflaged, small, incomplete, blurry, and noisy. The complexity of query images exceeds that of support images, providing a more realistic portrayal of real-world domain applications.
Then, evaluation experiments on this benchmark reveal that existing methods exhibit significantly poorer performance on challenging query images compared to simpler ones. The primary issue lies in the inadequate utilization of crucial category details from support images to effectively represent query features, making it challenging to eliminate distractions from complex backgrounds or noise.

To tackle this issue, a novel conditional representation learning network (CRLNet) is introduced, as illustrated in Figure~\ref{Fig: comparison}. CRLNet comprises three key components: a feature extractor, a conditional learner, and a re-representation learner. Additionally, contrastive learning loss is integrated to guide the training of CRLNet.
The core concept is to interact with the important information in the support and query to generate a similarity relationship matrix among them. By using their respective relationship matrices, it captures the fine-grained class-distinguishing features in the support or query to obtain feature representations with more class information. For difficult image queries, this representation method uses the important information from the support as a condition to guide the model in removing complex background features from the query images, thereby enhancing the model's performance consistency in complex environments. By integrating a contrastive learning framework, CRLNet effectively minimizes intra-class differences while maximizing inter-class differences at the feature representation level.

Finally, the comparative experimental results showcase the superiority of CRLNet, with ablation experiments confirming the effectiveness of each module within the network. The experimental outcomes demonstrate that CRLNet surpasses existing methods, and individual module tests within CRLNet validate their efficacy.

In summary, this paper has the following contributions:
\begin{itemize}
\item This paper introduces the concept of ``environmental robustness'' in few-shot learning, aiming to improve the performance consistency of few-shot learning methods in complex and diverse real-world environments. To evaluate this, a new real-world multi-domain few-shot learning benchmark is established, consisting of six datasets across four domains, with images fully manually annotated for environmental difficulty.
\item A novel conditional representation learning network is introduced to enhance the representation generalization ability from simple images to difficult ones. The aim is to mutually constrain the representation processes of known and unknown images, reducing intra-class distances and increasing inter-class distances at the feature level.
\item Comparative experiments demonstrate that CRLNet consistently outperforms existing SOTA methods across all evaluation datasets in the RD-FSL benchmark, with improvements ranging from 6.83\% to 16.98\%, highlighting its superiority and generalizability.
\end{itemize}

\section{Related Works}
\label{sec:related}

\subsection{Few-Shot Learning}

{\color{recolor}Few-shot learning (FSL) aims to acquire generalized meta-abilities or meta-knowledge from a large volume of readily available base class data. When faced with novel classes in a specific application domain, FSL facilitates rapid learning using only a small amount of support data to accurately predict unknown query data.}
FSL methods are typically classified into three categories: meta-learning-based, metric learning-based, and data augmentation-based. 
Meta-learning-based methods~\cite{FinnAL17, FallahMO21, QinSJ23} focus on training models across diverse tasks to acquire a ``learning to learn'' ability, enabling models to quickly learn new category knowledge from support images.
Metric learning-based methods~\cite{ZhangCLS23, XieLLWL22, ZhangLYCCC24} aim to learn a metric function to map similar images to higher similarity and dissimilar ones to lower similarity. 
Data augmentation-based methods~\cite{YangWLX22, XuLHAS21, Song0CMS23} focus on augmenting a large amount of image or feature data to alleviate severe overfitting issues caused by the limited number of support images.

Recently, researchers have increasingly focused on applying FSL to various application areas such as medical cell classification~\cite{GuoDJGTWZ23}, industrial defect identification~\cite{ZhouLZWC23}, leaf disease classification~\cite{GargS23a, BeiCZHZ24}. The semantic gap between images in these domains and base class images poses a challenge to the generalization performance of existing FSL methods. Furthermore, when faced with more complex images in real-world scenarios, current FSL based on pre-trained models struggles to quickly grasp the distribution knowledge of new classes accurately from a small support set and adapt to more challenging query images. Therefore, this paper aims to introduce a novel approach from the perspective of feature representation to enhance the performance of FSL in domains, particularly when handling difficult samples.

\subsection{Related Datasets}
\label{subsec:related}
In the early stages, FSL was primarily evaluated on general datasets, including miniImageNet, tieredImageNet, and CIFAR~\cite{VinyalsBLKW16, GuoGWFYZG23}, subsets derived from the ImageNet dataset. These datasets are coarse-grained. Subsequently, more works introduced a plethora of fine-grained evaluation datasets such as CUB, Stanford Cars, and Stanford Dogs~\cite{ZhaTST23, HuangZZXW21, TianX23}. 
Later on, to assess the performance of FSL methods in application domains, Guo Y et al.~\cite{GuoCKCSSRF20} introduced a cross-domain few-shot learning (CD-FSL) benchmark. This benchmark includes two medical domain datasets (ChestX and ISIC), as well as an agricultural domain dataset and a satellite image dataset. They evaluated previous FSL methods and found that when faced with significant semantic or modal differences in validation datasets, FSL methods generally performed poorly, struggling to learn high generalization meta-knowledge.
Furthermore, Fereshteh.S et al.~\cite{abs-2206-00092} introduce the few-shot classification of histological images (FHIST) benchmark.

However, when applying FSL to domain visual recognition, not only do researchers encounter semantic gaps when transitioning from general datasets to specific domains, but they also face challenges presented by a plethora of distracting backgrounds or unstable shooting conditions in real-world scenarios, leading to target difficulties or image blurriness.
Regrettably, existing datasets have overlooked the crucial impact of environmental robustness in practical applications. To address this, this paper proposes the RD-FSL benchmark, aiming to enhance the performance and robustness of FSL in real-world applications.

\subsection{Contrastive Learning}
\label{subsec:con}
Recently, contrastive learning~\cite{PengL0LZ0H23, ChenK0H20, HeCXLDG22}, utilizing instance discrimination as a pretext task, has emerged as a leading method in self-supervised representation learning. 
The training process of contrastive learning structures the training data into positive or negative pairs and employs a loss function that reduces the distance between positive pairs while increasing the distance between negative pairs to optimize the model. 
Early methods that introduced contrastive learning into FSL were primarily focused on incorporating self-supervised pretext tasks such as rotation~\cite{GidarisSK18} and jigsaw~\cite{NorooziF16} as auxiliary losses~\cite{GidarisBKPC19, ChenGZHW21}. In recent years, the focus has shifted towards directly integrating contrastive learning of instance discrimination into the pre-training of FSL~\cite{LiuF0YLWZ21, MaXHCGA21, DoerschGZ20, YangWZ22}. 


These methods primarily focus on leveraging the known category information in few-shot learning tasks to design new optimization functions by integrating contrastive learning losses. This aims to combine the representational strengths of unsupervised contrastive learning with the characteristics of known classes in few-shot learning tasks. {\color{recolor}However, data from real-world environments often deviates significantly from the training data distribution. As a result, these methods struggle to filter out noisy features in challenging images, frequently misaligning features from non-target classes and leading to biased feature representations for difficult images.}
To address this, the paper proposes conditional representation learning. It focuses on how to accurately capture the critical category information of known and unknown images, even when unknown images contain a lot of noise. 

\section{The Proposed Benchmark}
\label{sec:proposed}

\begin{figure*}[t]
	\centering
	\includegraphics[width=1.0\linewidth]{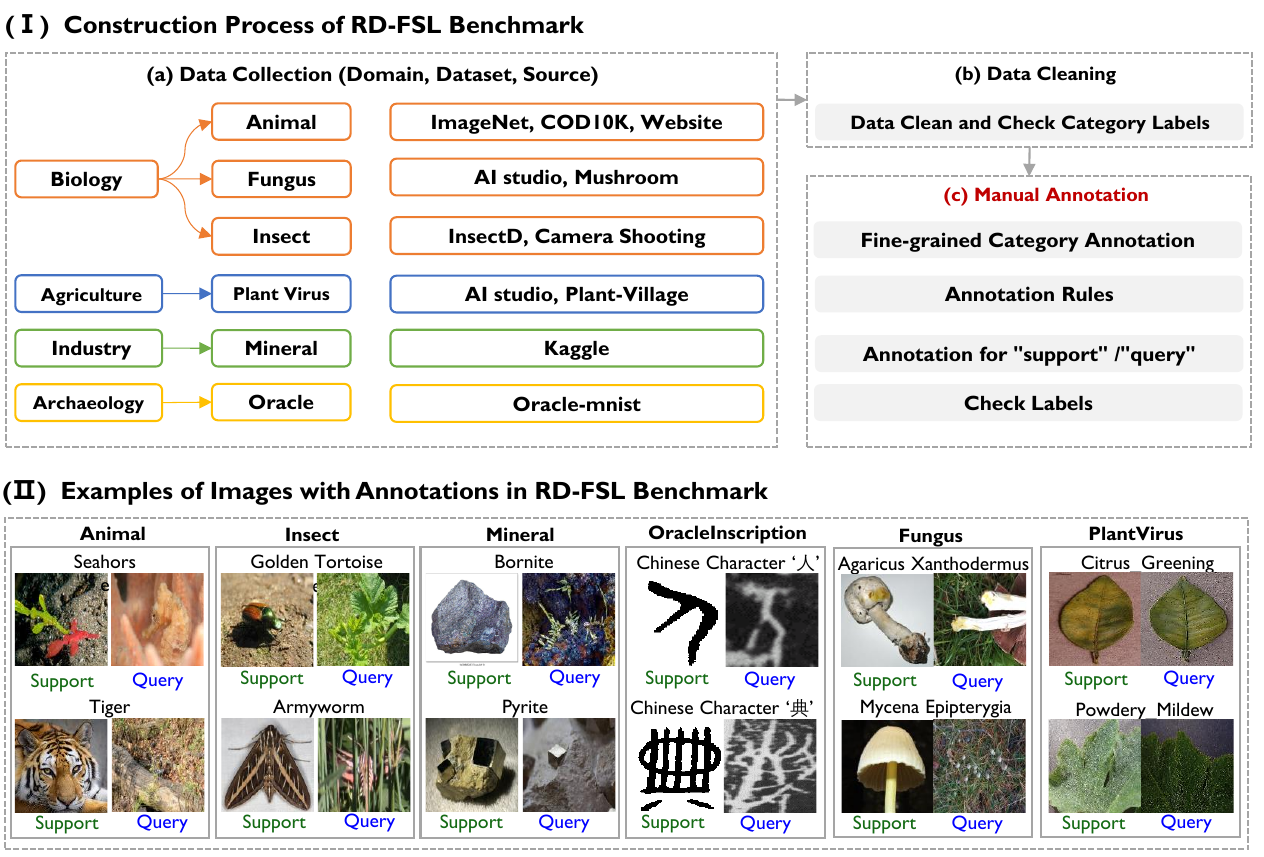}
	\caption{(\uppercase\expandafter{\romannumeral1}) The construction process of the real-world multi-domain few-shot visual recognition (RD-FSL) benchmark includes three steps: data collection, data cleaning, and manual annotation. The manual annotation step involves human labeling of fine-grained classification labels and difficulty (support/query) labels. Additionally, (\uppercase\expandafter{\romannumeral2}) examples of images and annotations within it demonstrate the contrast in difficulty levels for the same category labeled as support and query across six validation datasets.}
	\label{Fig: Benchmark}
\end{figure*}

\subsection{Introduction of the Proposed Benchmark}
\label{subsec:introduction}
This paper addresses the shortage of evaluation datasets tailored for real-world domain applications by introducing a new RD-FSL benchmark. Unlike previous datasets, this benchmark highlights three key factors: 1) Broader Coverage: it includes a wider range of application domains by incorporating more datasets; 2) Annotation: each image is manually annotated for difficulty, with straightforward samples labeled as ``support'' and more complex ones as ``query''; and 3) Complexity: it integrates manually generated blurred images with diverse resolutions and noise levels.
\begin{table}[th]
	\centering
  	\caption{The statistics and categorization of each evaluation dataset in RD-FSL Benchmark.}
  \scalebox{1.0}{
	\begin{tabular}{c|c|c|c|c}
\hline
\hline
{Dataset} &{Domain} &{Class(\#)} &Num(\#) &Grained\\
\hline
Animal &Biology &$34$ &$50,304$&Coarse\\
Insect &Biology &$70$ &$22,242$&Fine\\
Fungus &Biology &$51$ &$21,096$&Fine\\
Mineral &Mining industry &$6$ &$867$&Fine\\
OracleIn &Archaeology &$241$ &$308,593$&Fine\\
PlantVirus &Agriculture &$6$ &$1,810$&Fine\\
\hline
\hline
	\end{tabular}}
	\label{tab:benchmark}
\end{table}

Table~\ref{tab:benchmark} presents the details of the six datasets. The Animal dataset includes $34$ categories, such as ``Dog'' and ``Snake'', and is classified as coarse-grained. In contrast, the Insect and Fungus datasets include $70$ and $51$ species, respectively, with subtle differences, classifying them as fine-grained. The Mineral dataset contains 6 unique mineral types, placing it in the fine-grained category. The OracleInscription Recognition dataset offers $308,593$ images across $241$ Chinese character categories. The PlantVirus dataset, relevant to agriculture, involves identifying $6$ disease types based on subtle plant lesions. Additionally, over $400,000$ images in the RD-FSL benchmark are manually annotated for difficulty levels.

\subsection{Construction Process and Annotation Rules}
\label{subsec:construction}
As in Figure~\ref{Fig: Benchmark} (\uppercase\expandafter{\romannumeral1}), the construction process comprises three primary stages: data collection, data cleaning, and manual annotation. First, we collected images from established datasets including ImageNet~\cite{DengDSLL009}, COD10K~\cite{FanJSCS020}, Kaggle~\cite{kaggle}, AI studio~\cite{AIstudio}, Mushroom~\cite{AnagnostopoulouREFM23}, Insect Classification~\cite{WuZLCY19}, Plant-Village~\cite{HughesS15}, and Oracle-mnist~\cite{abs-2205-09442}. Additionally, we acquired some images from the internet and captured new ones to ensure that each category's image quantity meets the validation criteria. Second, we meticulously assessed each image, eliminating duplicates or low-quality images, and manually confirmed the accuracy of existing category labels.
Third, we manually labeled images lacking category tags, with a particular focus on fine-grained category labels. Besides, each image is categorized as ``support'' or ``query'' and evaluated by a minimum of three individuals. The final label is determined based on the most frequently assigned tag. The specific rules for ``query'' images are outlined as follows: 
\begin{itemize} 
\item Camouflaged: When the subject and background exhibit similar attributes like color and texture. 
\item Small: Targets occupy less than approximately 1\% of the total pixels in the image. 
\item Incomplete: When more than 50\% of the crucial attributes in the target are absent. 
\item Blurry: A minimum of 5\% of images in each dataset are randomly altemycolor to create images with diverse resolutions or varying noise levels.
\end{itemize}

From Figure~\ref{Fig: Benchmark} (\uppercase\expandafter{\romannumeral2}), it's clear that support images have sharp resolution and clear objects, while query images mimic those easily disturbed by complex backgrounds or noise in real-world settings. Current FSL models find it difficult to extend category knowledge to challenging query images during rapid learning on support images, resulting in prediction errors. In summary, the RD-FSL benchmark aims to provide a multi-domain application evaluation dataset that is more closely aligned with real-world scenarios.

\section{The Proposed Method}
\label{sec:method}

\begin{figure*}[t]
	\centering
	\includegraphics[width=1.0\linewidth]{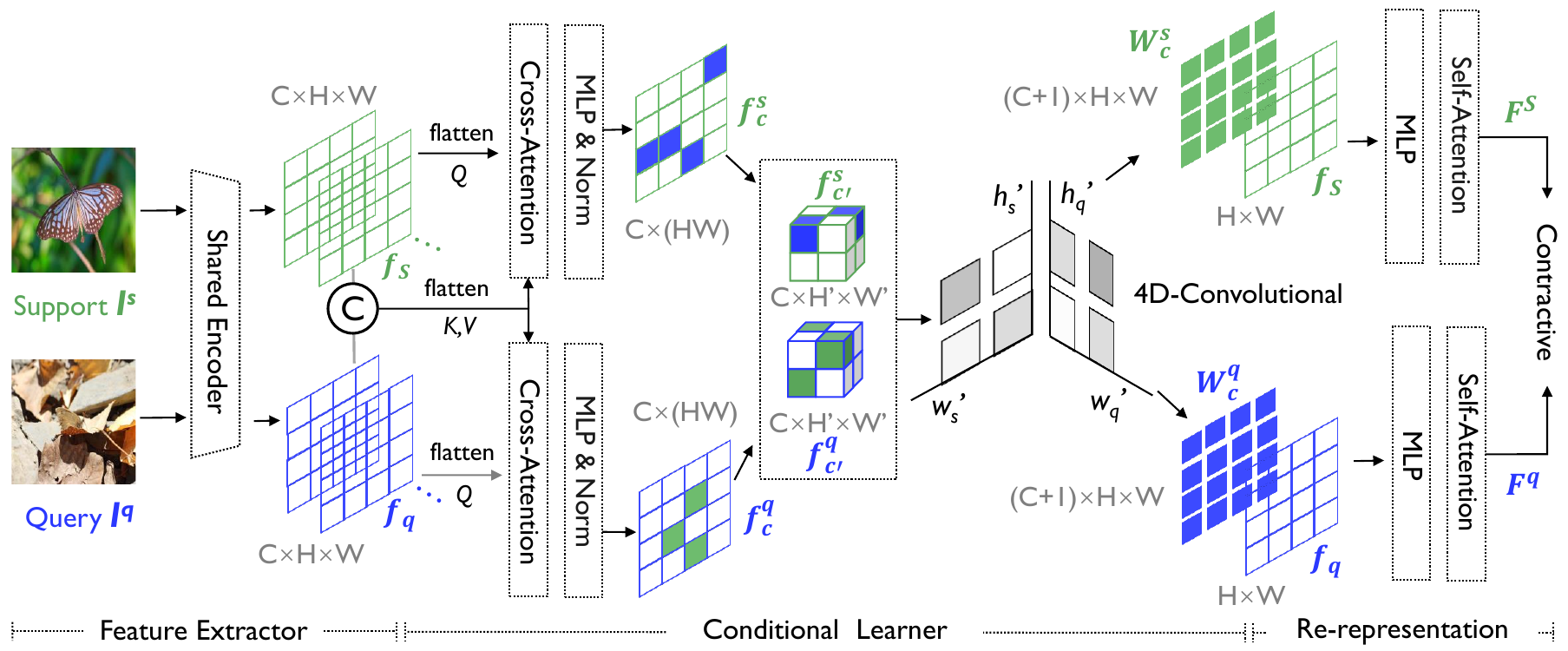}
	\caption{{\color{mycolor}The overview of the proposed conditional representation learning network (CRLNet) includes a feature extractor, a conditional learner, and a re-representation learner. The feature extractor maps support $\mathcal{I}^{s}$ and query images $\mathcal{I}^{q}$ to prototype feature matrices ${f}^{s}$ and ${q}^{s}$. Subsequently, the conditional learner learns the conditional matrices $\omega^{s}_{c}$ and $\omega^{q}_{c}$. Finally, the re-representation learner relearns the ${f}^{s}$ and ${q}^{s}$ with $\omega^{s}_{c}$ and $\omega^{q}_{c}$ to obtain the final feature representation $\mathcal{F}^{s}$ and $\mathcal{F}^{q}$. The entire CRLNet is supervised with a contrastive learning loss.}}
	\label{Fig: Method}
\end{figure*}

\subsection{Problem Definition}
\label{sec:Problem}
The model is initially trained on a large set of images, $\mathcal{T}_{base}$, from base classes $\mathcal{C}_{base}$ and then evaluated on images $\mathcal{T}_{novel}$ from novel classes $\mathcal{C}_{novel}$ within a specific domain, ensuring $\mathcal{C}_{base} \cap \mathcal{C}_{novel} = \emptyset$. {\color{recolor}The evaluation employs a meta-test approach, involving support set $\mathcal{S}_{i}$ for the model to quickly learn new class information and a query set $\mathcal{Q}_{i}$ for testing.} The support set is organized in an $N$-way-$K$-shot format, where $N$ is the number of novel classes per evaluation, and $K$ is the number of labeled images provided for each class. During each evaluation, the pre-trained model must rapidly acquire new category information from $\mathcal{S}_{i}$ to assist in predicting $\mathcal{Q}_{i}$.  
For fair comparison and optimal model performance, our experiments differ from other benchmarks in two key aspects. First, we use ILSVRC~\cite{DengDSLL009} as $\mathcal{T}_{base}$ instead of the miniImageNet dataset used in previous studies. Second, $\mathcal{S}_{i}$ and $\mathcal{Q}_{i}$ are extracted from data labeled as ``support'' or ``query'' in the RD-FSL benchmark dataset.

\subsection{Overview}
This paper introduces CRLNet, designed to integrate key features of $\mathcal{S}_{i}$ and $\mathcal{Q}_{i}$ from $\mathcal{C}_{novel}$ into each other's re-representation process. This integration aims to reduce category distances at the feature level while enhancing inter-class differences. An overview of CRLNet is shown in Figure~\ref{Fig: Method}. CRLNet utilizes a Siamese structure and comprises three modules: a feature extractor, a conditional learner, and a re-representation learner.

The feature extractor derives prototype features ${f}_{s}$ and ${f}_{q}$ for $\mathcal{S}_{i}$ and $\mathcal{Q}_{i}$, respectively. The conditional learner comprises a cross-attention layer and a 4D convolution.   
{\color{mycolor}The cross-attention mechanism generates interacted support and query features, while the 4D convolution produces feature weight matrices $\mathcal{W}^{s}_{c}$ and $\mathcal{W}^{q}_{c}$. This 4D convolution structure creates an uncompressed combination of support and query information. By performing convolutional calculations in various directions, it integrates crucial information from the support or query into the other. The sliding window approach used in convolutional calculations ensures uniform computation of relevant information in the neighborhood, aiding in capturing fine-grained relevant features and mitigating the issue of noisy features disrupting challenging query images.}
Finally, the re-representation learner, consisting of self-attention and an MLP, readapts the combination of weight matrices and prototype feature information through self-adaptive learning.  

When training, we select an equal number of same-class and different-class image pairs in each batch to prevent training instability due to imbalanced positive and negative samples. In testing, all support images and the query image are sequentially input into CRLNet to generate multiple sets of re-represented features. We then train the classifier online and predict based on the average of all generated query features.

\subsection{Feature Extractor and Conditional Learner}
Each time, we randomly select two images from $\mathcal{T}_{base}$ to serve as $\mathcal{S}_{i}$ and $\mathcal{Q}_{i}$. If the labels $\mathcal{L}_{s}$ and $\mathcal{L}_{q}$ of these images match, they are considered a positive pair; otherwise, they form a negative pair. {\color{recolor}After being processed by the feature extractor $\mathcal{E}(x)$, the corresponding prototype feature matrices are obtained as ${f}^{s}_{i} \in \mathbb{R}^{W \times H \times C}$ and ${f}^{q}_{i} \in \mathbb{R}^{W \times H \times C}$.}
In the experiments, we employed backbone networks with the same configurations as other methods for feature extraction, including ResNet-12, ResNet-50, and ViT. The prototype features represent the entire image's feature representation as captured by the backbone network. However, when handling query images in complex environments, the backbone network may be overwhelmed by numerous complex noise and background features, hindering its ability to focus on essential category information. To address this, CRLNet incorporates a conditional learner designed to capture the similarity between important regions of the support and query images through their interaction. This process generates a key information matrix for the support based on the query, or vice versa.

In the conditional learner, a cross-attention module and a 4D convolution module are designed. The cross-attention module is intended to compute the similarity of the concatenated prototype features of the support and query from a global perspective, thereby obtaining support and query feature vectors that incorporate global information.
Specifically, after obtaining the prototype feature matrices ${f}^{s}_{i} \in \mathbb{R}^{W \times H \times C}$ and ${f}^{q}_{i} \in \mathbb{R}^{W \times H \times C}$, we first flatten ${f}^{s}_{i}$ and ${f}^{q}_{i}$ to obtain the prototype features in $\mathbb{R}^{(W \times H) \times C}$. We then concatenate ${f}^{s}_{i}$ and ${f}^{q}_{i}$ to form the aggregated prototype feature ${f}^{m}_{i} \in \mathbb{R}^{W \times H \times 2C}$, which becomes $\mathbb{R}^{(2W \times H) \times C}$ after flattening. We use cross-attention mechanisms to calculate the similarity matrix between ${f}^{s}_{i}$ or ${f}^{q}_{i}$ and ${f}^{m}_{i}$. Previous work mostly computed the similarity between ${f}^{s}_{i}$ and ${f}^{q}_{i}$ directly as their similarity matrix, which might overly emphasize the key information of one party while neglecting its characteristics. 

Then, we propose an operation to calculate the similarity between prototype features and aggregated features. Specifically, we utilize the scaled dot-product attention, as shown in Eq.~(\ref{attention}), to compute the similarity matrix between ${f}^{s}_{i}$ or ${f}^{q}_{i}$ with ${f}^{m}_{i}$:
\begin{equation}
\label{attention}
\textit{Attention}(Q,K,V) = \text{softmax}\left(\frac{{QK^{T}}}{\sqrt{d}}\right){V},
\end{equation}
where $Q$, $K$, and $V$ are sets of query, key, and value vectors packed into matrices, and $d$ is the dimension of the query and key vectors. We consider ${f}^{s}_{i}$ or ${f}^{q}_{i}$ as the $Q$ value, and the aggregated feature ${f}^{m}_{i}$ as the $K$ and $V$ values. Initially, we compute dense pixel-wise attention between ${f}^{s}_{i}$ or ${f}^{q}_{i}$ and ${f}^{m}_{i}$ to obtain a similarity matrix. This weight matrix is then multiplied with ${f}^{m}_{i}$ to identify the most relevant feature matrix in ${f}^{s}_{i}$ or ${f}^{q}_{i}$ associated with ${f}^{m}_{i}$:
\begin{equation}
\label{cross}
f_{ic} = \textit{Correlation}(f_{i}, {f}^{m}_{i}) = \textit{softmax}\left(\frac{{f_{i}{f}^{m}_{iT}}}{\sqrt{d}}\right){f}^{m}_{i}.
\end{equation}
In our computations, we adhere to the original Transformer approach~\cite{VelickovicCCRLB18}, using sine and cosine functions of different frequencies for positional encoding. Calculating the similarity between prototype features and aggregated features can mitigate the excessive representation of the direct interaction of two prototype features in previous works, which is essential for conveying conditional information in subsequent steps.

After that, the 4D convolution module performs convolution calculations on the integrated support and query features from different directions. This process yields a mapping relation matrix of important information in the support relative to the global features of the query, as well as a mapping relation matrix of important information in the query based on the global information of the support.
After obtaining $f^{s}_{ic} \in \mathbb{R}^{(W \times H) \times C}$ and $f^{q}_{ic} \in \mathbb{R}^{(W \times H) \times C}$, we perform a permutation operation to obtain $f^{s}_{ic'} \in \mathbb{R}^{W^{s} \times H^{s} \times C}$ and $f^{q}_{ic'} \in \mathbb{R}^{W^{q} \times H^{q} \times C}$. This is done to maintain the same dimensions as the prototype features, which is beneficial for subsequent adaptive transfer learning. By multiplying $f^{s}_{ic'}$ and $f^{q}_{ic'}$, i.e., $f^{m}_{c'} = f^{s}_{c'} \otimes f^{q}_{c'}$, we obtain the aggregated relationship matrix $f^{m}_{ic'} \in \mathbb{R}^{W^{s} \times H^{s} \times W^{q} \times H^{q} \times C}$, which contains the complete interaction information between support and query features.   
The aggregated relationship matrix is not compressed and does not require alignment. This is to maximally preserve the complete information in the support and query, preventing the loss of critical features in challenging images during the filtering process.

{\color{recolor}Next,} we design bidirectional 4D convolutions for support and query, where $K \times L \times M \times N$ is the shape of the 4D convolutional kernel. The basic 4D convolution operation is expressed by the following formula~\cite{ZhangGHS020}:
\begin{equation}
\begin{aligned}
\label{4D}
y_{k,l,m,n} = &F_{C}(\sum_{c}^{C}\sum_{w^{s}=0}^{W^{s}}\sum_{h^{s}=0}^{H^{s}}\sum_{w^{q}=0}^{W^{q}}\sum_{h^{q}=0}^{H^{q}}W_{(h^{q},w^{q},h^{s},w^{s})}\\
&x_{c,(k+h^{q}),(l+w^{q}),(m+w^{s}),(n+w^{s})})+b_{h^{q},w^{q},h^{s},w^{s}}{\color{recolor}.}
\end{aligned}
\end{equation}
In CRLNet, assuming $W_{j}^{c}$ is the weight at position $k, l, m, n$ of the kernel, we set $m$ and $n$ to $1$ for support and $k$ and $l$ to $1$ for the query. Learning from the same convolutional kernel in different directions helps in conveying more important information for support and query. The corresponding conditional relationship matrices $\omega^{s}_{c} \in \mathbb{R}^{W \times H}$ and $\omega^{q}_{c} \in \mathbb{R}^{W \times H}$ are obtained using the following formula:
\begin{equation}
\begin{aligned}
\label{4D_support}
\omega^{s}_{c} = &F_{Cs}(\sum_{c}^{C}\sum_{w^{s}=0}^{W^{s}}\sum_{h^{s}=0}^{H^{s}}\sum_{w^{q}=0}^{W^{q}}\sum_{h^{q}=0}^{H^{q}}W_{(h^{q},w^{q},1,1)}\\
&x_{c,(k+h^{q}),(l+w^{q}),(1+{\color{mycolor}h}^{s}),(1+w^{s})})+b_{h^{q},w^{q},h^{s},w^{s}},
\end{aligned}
\end{equation}
\begin{equation}
\begin{aligned}
\label{4D_query}
\omega^{q}_{c} = &F_{Cs}(\sum_{c}^{C}\sum_{w^{s}=0}^{W^{s}}\sum_{h^{s}=0}^{H^{s}}\sum_{w^{q}=0}^{W^{q}}\sum_{h^{q}=0}^{H^{q}}W_{(1,1,h^{s},w^{s})}\\
&x_{c,(1+h^{q}),(1+w^{q}),(m+{\color{mycolor}h}^{s}),(n+w^{s})})+b_{h^{q},w^{q},h^{s},w^{s}},
\end{aligned}
\end{equation}
where $F_{C}(.)$ is an activation function and $b_{h^{q},w^{q},h^{s},w^{s}}$ is the bias of the computed feature map.  

To explain in detail, when \( m \) and \( n \) are set to 1, the convolutional kernel aggregates the support feature matrix in the first two dimensions through convolution computation, compressing it into the query feature matrix. This convolution computation sequentially calculates the similarity relationship matrix between important information in the global features of the support and local features in the query. The characteristic of convolution is to use a sliding window approach to compute this relationship matrix, allowing the local correlation matrix to be continuously mapped in the subsequent results, maintaining the consistency of this relationship matrix. Conversely, the same applies to the support features. Therefore, compared to the previous step of cross-attention, this special 4D convolution design can more precisely capture the similarity between important regions of the support and query without compressing or manually aligning the support and query features. It also maintains the continuity of the neighborhood, making the resulting relationship matrices \(\omega^{s}_{c}\) and \(\omega^{q}_{c}\) more accurate, global, and consistent.

\subsection{Re-representation Learner}
After obtaining the conditional matrices $\omega^{s}_{c} \in \mathbb{R}^{W \times H}$ and $\omega^{q}_{c} \in \mathbb{R}^{W \times H}$, we aggregate them with the prototype matrices $f^{s}\in \mathbb{R}^{W \times H \times C}$ and $f^{q}\in \mathbb{R}^{W \times H \times C}$ respectively for re-representation. The feature dimension after concatenation is $\mathbb{R}^{W \times H \times (C+1)}$. We introduce an MLP layer and a self-attention layer to transfer crucial information from the conditional matrices interacting with the other features to the prototype matrices while ensuring the preservation of their key features. The MLP layer compresses the input dimensions from $\mathbb{R}^{W \times H \times (C+1)}$ to $\mathbb{R}^{(W \times H) \times C}$, aiming to adaptively learn the fusion of conditional matrices and prototype features. Subsequently, this feature passes through the self-attention layer and is mapped to the final support feature vector $F^{s} \in \mathbb{R}^{C}$ and query feature vector $F^{q} \in \mathbb{R}^{C}$. The specific operations are as follows:
\begin{equation}\label{self}
\hat{f'} = \textit{Attention}(f'W^{Q}_{\phi},f'W^{K}_{\phi},f'W^{V}_{\phi}), \hat{f'} \in \mathbb{R}^{(W \times H)\times C},
\end{equation}
\begin{equation}
\label{MLP}
F_{i} = \textit{MLP}(\textit{LN}(f'+\hat{f'})).
\end{equation}

\subsection{Loss Function}
After the above operations, we have obtained the re-representation features based on the conditional relationship matrices $F^{s}$ and $F^{q}$. To bring similar features closer and push dissimilar features apart at the representation level, we utilize contrastive learning loss for optimization. It is worth noting that existing methods often divide training into pre-training and meta-training steps, while CRLNet trains in a single step. This means that it supervises the overall network, including the feature extraction backbone, using the contrastive learning loss for training.  
In a batch size, to balance the number of positive and negative samples and ensure smooth model training, we set half of the sample pairs to be of the same class. Therefore, for individual $F^{s}$ and $F^{q}$, we calculate their difference using the L2 distance:
\begin{equation}
\label{L2}
d_{(F^{q},{F^{s}})} = \Vert F^{s} - F^{q} \Vert^{2}.
\end{equation}
Based on $d_{(F^{q},{F^{s}})}$, the total loss in one batch task can be calculated as:
\begin{equation}
\label{loss}
\mathcal{L} = -\frac{1}{N}\sum^{N}_{i=1}\textbf{I}(\mathcal{L}_{q}==\mathcal{L}_{s})log(d_{(F^{q},{F^{s}})}),
\end{equation}
where $\textbf{I}(\mathcal{L}_{q}==\mathcal{L}_{s})$ equals $1$ when the label of query $\mathcal{L}_{q}$ and support $\mathcal{L}_{s}$ are equal, otherwise 0. $N$ is the total number of extracted positive and negative samples, and $i$ is the number of extraction iteration.

\section{Experiments}

\begin{table*}[th]
\renewcommand{\arraystretch}{1.0}
	\centering
  	\caption{The comparison experiment results between CRLNet and SOTA methods in 5-way-1-shot, 5-way-5-shot, and 5-way-10-shot settings on two datasets in the field of biology (Animal and Insect). The numbers in \textbf{Bold} indicate the best performance, while the \underline{underline} one {\color{recolor}denotes} the second best. $^\ast$ indicates that these methods use a transductive setting, while other methods use an inductive setting.}
  \scalebox{1.0}{
	\begin{tabular}{l|c|ccc|ccc}
\hline
\hline
Method &{Backbone} &\multicolumn{3}{c|}{{Animal}} &\multicolumn{3}{c}{{Insect}}\\
&&1-shot &5-shot &10-shot &1-shot &5-shot &10-shot\\
\hline

MAML{~\color{gray}{[ICML2017]}}~\cite{FinnAL17} &ResNet-12&$28.06_{\pm0.55}$&$29.01_{\pm0.53}$&$31.22_{\pm0.53}$
&$27.83_{\pm0.51}$&$29.88_{\pm0.47}$&$31.78_{\pm0.48}$\\

ProtoNet{~\color{gray}{[NIPS2017]}}~\cite{SnellSZ17}  &ResNet-12&$30.67_{\pm0.58}$&$39.92_{\pm0.81}$&$43.02_{\pm0.82}$
&$29.15_{\pm0.50}$&$44.28_{\pm0.71}$&$47.48_{\pm0.68}$\\

RelationNet{~\color{gray}{[CVPR2020]}}~\cite{ZhuangTYMZJX18}  &ResNet-12&$30.72_{\pm0.57}$&$38.33_{\pm0.70}$&$40.14_{\pm0.70}$
&$29.82_{\pm0.50}$&$41.77_{\pm0.63}$&$44.09_{\pm0.63}$\\

IE{~\color{gray}{[CVPR2021]}}~\cite{Rizve0KS21}  &ResNet-12&$30.84_{\pm0.60}$&$41.88_{\pm0.71}$&$46.17_{\pm0.68}$&$34.85_{\pm0.53}$&$44.58_{\pm0.59}$&$49.14_{\pm0.58}$\\

DeepDBC{~\color{gray}{[CVPR2022]}}~\cite{XieLLWL22}  &ResNet-12&$32.78_{\pm0.62}$&$41.37_{\pm0.86}$&$45.07_{\pm0.85}$
&$28.54_{\pm0.47}$&$42.24_{\pm0.78}$&$45.53_{\pm0.78}$\\

DeepEMD{~\color{gray}{[TPAMI2023]}}~\cite{ZhangCLS23}  &ResNet-12&$34.54_{\pm0.63}$&$42.24_{\pm0.82}$&$45.10_{\pm0.80}$
&$35.78_{\pm0.63}$&$\underline{49.52_{\pm0.75}}$&$\underline{54.85_{\pm0.68}}$\\

RanKDNN{~\color{gray}{[AAAI2023]}}~\cite{GuoGWFYZG23}  &ResNet-12&$\underline{35.03_{\pm0.50}}$&$40.92_{\pm0.25}$&$42.15_{\pm0.60}$&$\underline{37.99_{\pm0.60}}$&$46.53_{\pm0.53}$&$50.26_{\pm0.35}$\\

ESPT{~\color{gray}{[AAAI2023]}}~\cite{RongLSCX23} 
 &ResNet-12&$34.12_{\pm0.65}$&$\underline{43.92_{\pm0.66}}$&$\underline{47.37_{\pm0.68}}$
&$33.77_{\pm0.58}$&$45.00_{\pm0.59}$&$48.88_{\pm0.56}$\\
{\color{mycolor}TPN$^\ast$}{~\color{gray}{[ICLR2019]}}~\cite{LiuLPKYHY19} &ResNet-12 &$26.66_{\pm0.55}$&$34.82_{\pm0.50}$&$36.67_{\pm0.59}$&$28.08_{\pm0.47}$&$45.51_{\pm0.56}$&$45.41_{\pm0.59}$\\
{\color{mycolor}TIM+PR$^\ast$}{~\color{gray}{[ICML2021]}}~\cite{CuiG21} &ResNet-12 &$27.51_{\pm0.95}$&$33.28_{\pm0.67}$&$37.24_{\pm0.39}$&$32.32_{\pm0.94}$&$46.00_{\pm0.64}$&$48.25_{\pm0.83}$\\
{\color{mycolor}EASY$^\ast$}{~\color{gray}{[arxiv2022]}}~\cite{BendouHLLPPG22}&ResNet-12 &$30.67_{\pm0.58}$&$40.00_{\pm0.65}$&$44.00_{\pm0.34}$&$42.67_{\pm0.85}$&$43.33_{\pm0.58}$&$48.00_{\pm0.76}$\\
{\color{mycolor}protoLP$^\ast$}{~\color{gray}{[CVPR2023]}}~\cite{ZhuK23} &ResNet-12 &$36.35_{\pm0.64}$&$48.72_{\pm0.59}$&$53.05_{\pm0.47}$&$33.53_{\pm0.69}$&$45.75_{\pm0.72}$&$51.67_{\pm0.58}$\\
\textbf{CRLNet (Ours)} 
 &ResNet-12&\bm{$46.71_{\pm0.84}$}&\bm{$55.47_{\pm0.82}$}&\bm{$60.14_{\pm0.79}$}
&\bm{$54.08_{\pm0.81}$}&\bm{$64.89_{\pm0.76}$}&\bm{$69.38_{\pm0.72}$}\\
\hline
\hline

IE{~\color{gray}{[CVPR2021]}}~\cite{Rizve0KS21} 
&ResNet-50&$31.69_{\pm0.60}$&$38.89_{\pm0.60}$&$41.28_{\pm0.59}$
&\underline{$38.73_{\pm0.65}$}&\underline{$49.66_{\pm0.62}$}&$52.64_{\pm0.58}$
\\

DeepEMD{~\color{gray}{[TPAMI2023]}}~\cite{ZhangCLS23} 
&ResNet-50&$31.64_{\pm0.60}$&$38.42_{\pm0.54}$&$41.56_{\pm0.61}$
&$32.59_{\pm0.57}$&$43.72_{\pm0.57}$&$47.45_{\pm0.60}$\\

StyleAdv{~\color{gray}{[CVPR2023]}}~\cite{FuXFJ23} 
&ResNet-50&$32.08_{\pm0.64}$&$36.79_{\pm0.64}$&$39.60_{\pm0.63}$
&$33.12_{\pm0.66}$&$40.36_{\pm0.61}$&$42.34_{\pm0.60}$\\

ESPT{~\color{gray}{[AAAI2023]}}~\cite{RongLSCX23} 
&ResNet-50&\underline{$37.14_{\pm0.70}$}&$\underline{46.11_{\pm0.65}}$&\underline{$48.69_{\pm0.68}$}
&$36.86_{\pm0.62}$&$49.04_{\pm0.63}$&\underline{$52.70_{\pm0.62}$}\\

{\color{mycolor}protoLP$^\ast$}{~\color{gray}{[CVPR2023]}}~\cite{ZhuK23} &ResNet-50&$37.65_{\pm0.78}$&$47.45_{\pm 0.82}$&$51.66_{\pm 0.75}$&$42.53_{\pm 0.71}$&$51.76_{\pm0.70}$&$55.87_{\pm0.65}$\\
{\color{mycolor}FM$^\ast$}{~\color{gray}{[TIP2024]}}~\cite{WangLXHH24} &ResNet-50&$42.03_{\pm0.35}$&$50.37_{\pm 0.49}$&$52.75_{\pm 0.67}$&$44.27_{\pm 0.37}$&$51.90_{\pm0.61}$&$57.44_{\pm0.50}$\\

\textbf{CRLNet (Ours)} 
&ResNet-50&\bm{$62.76_{\pm0.81}$}&\bm{$70.79_{\pm0.69}$}&\bm{$72.49_{\pm0.66}$}
&\bm{$64.66_{\pm0.81}$}&\bm{$75.44_{\pm0.69}$}&\bm{$78.50_{\pm0.62}$}\\
\hline
\hline
FewTURE{~\color{gray}{[CVPR2020]}}~\cite{YeHZS20} &ViT&$34.28_{\pm0.50}$&$44.44_{\pm0.85}$&$47.13_{\pm0.82}$
&$32.59_{\pm0.50}$&$44.13_{\pm0.76}$&$49.14_{\pm0.76}$\\
HTCTrans{~\color{gray}{[CVPR2022]}}~\cite{HeLZZGYZ22} &ViT&$42.15_{\pm0.70}$&$47.82_{\pm0.85}$&$50.12_{\pm0.88}$
&\underline{$47.47_{\pm0.66}$}&$59.03_{\pm0.61}$&$62.93_{\pm0.59}$\\
CPEA{~\color{gray}{[ICCV2023]}}~\cite{HaoH0WT023} 
&ViT&\underline{$42.46_{\pm0.74}$}&\underline{$52.07_{\pm0.73}$}&\underline{$54.71_{\pm0.72}$}
&$44.54_{\pm0.70}$&\underline{$60.67_{\pm0.64}$}&\underline{$64.00_{\pm0.59}$}
\\
\textbf{CRLNet (Ours)} 
&ViT&\bm{$70.86_{\pm0.78}$}&\bm{$77.90_{\pm0.64}$}&\bm{$79.20_{\pm0.61}$}
&\bm{$74.71_{\pm0.76}$}&\bm{$85.19_{\pm0.50}$}&\bm{$87.05_{\pm0.45}$}\\
\hline
\hline
	\end{tabular}}
	\label{tab:comparison_A}
\end{table*}

\begin{table*}[th]
\renewcommand{\arraystretch}{1.0}
	\centering
  	\caption{The comparison experiment results between CRLNet and SOTA methods on 5-way-1-shot, 5-way-5-shot, and 5-way-10-shot settings on two datasets in the field of the mining industry (Mineral) and archaeology (OracleInscription). The numbers in \textbf{Bold}  indicate the best performance, while the \underline{underline} one denotes the second best.}
  \scalebox{1.0}{
	\begin{tabular}{l|c|ccc|ccc}
\hline
\hline
Method &Backbone  &\multicolumn{3}{c|}{{Mineral}} &\multicolumn{3}{c}{{OracleInscription}}\\
&&1-shot &5-shot &10-shot &1-shot &5-shot &10-shot\\
\hline
MAML{~\color{gray}{[ICML2017]}}~\cite{FinnAL17} &ResNet-12&$37.30_{\pm0.55}$&$37.91_{\pm0.56}$&$38.96_{\pm0.52}$
&$25.67_{\pm0.58}$&$26.95_{\pm0.55}$&$28.23_{\pm0.50}$\\

ProtoNet{~\color{gray}{[NIPS2017]}}~\cite{SnellSZ17} &ResNet-12&$34.62_{\pm0.54}$&$52.29_{\pm0.69}$&$57.21_{\pm0.62}$
&\underline{$29.17_{\pm0.58}$}&$36.08_{\pm0.67}$&$37.85_{\pm0.69}$\\

RelationNet{~\color{gray}{[CVPR2020]}}~\cite{ZhuangTYMZJX18} &ResNet-12&$37.86_{\pm0.57}$&$47.40_{\pm0.67}$&$49.90_{\pm0.64}$
&$26.71_{\pm0.56}$&$32.16_{\pm0.61}$&$34.45_{\pm0.59}$\\

IE{~\color{gray}{[CVPR2021]}}~\cite{Rizve0KS21} &ResNet-12&$37.22_{\pm0.59}$&$51.51_{\pm0.57}$&$58.06_{\pm0.59}$&$25.96_{\pm0.56}$&$30.84_{\pm0.65}$&\underline{$37.87_{\pm0.68}$}\\

DeepDBC{~\color{gray}{[CVPR2022]}}~\cite{XieLLWL22} &ResNet-12&\bm{$43.64_{\pm0.61}$}&$53.98_{\pm0.90}$&$59.65_{\pm0.87}$
&$26.77_{\pm0.55}$&$25.46_{\pm0.59}$&$31.27_{\pm0.66}$\\

DeepEMD{~\color{gray}{[TPAMI2023]}}~\cite{ZhangCLS23} &ResNet-12&$39.62_{\pm0.63}$&\underline{$56.28_{\pm0.88}$}&\underline{$62.60_{\pm0.85}$}
&$27.85_{\pm0.60}$&\underline{$36.88_{\pm0.69}$}&$41.33_{\pm0.71}$\\

RanKDNN{~\color{gray}{[AAAI2023]}}~\cite{GuoGWFYZG23} &ResNet-12&$36.01_{\pm0.25}$&$55.98_{\pm0.54}$&$60.70_{\pm0.63}$&$28.20_{\pm0.50}$&$31.44_{\pm0.37}$&$35.80_{\pm0.36}$\\

ESPT{~\color{gray}{[AAAI2023]}}~\cite{RongLSCX23} 
&ResNet-12&\underline{$42.32_{\pm0.64}$}&$55.83_{\pm0.58}$&$60.96_{\pm0.55}$
&$28.33_{\pm0.59}$&$35.00_{\pm0.66}$&$36.99_{\pm0.65}$\\
\textbf{CRLNet (Ours)} 
&ResNet-12&$40.05_{\pm0.63}$&\bm{$57.37_{\pm0.62}$}&\bm{$65.61_{\pm0.58}$}
&\bm{$29.34_{\pm0.60}$}&\bm{$38.52_{\pm0.66}$}&\bm{$42.68_{\pm0.72}$}\\
\hline
\hline

IE{~\color{gray}{[CVPR2021]}}~\cite{Rizve0KS21} 
&ResNet-50&$29.97_{\pm0.59}$&$40.07_{\pm0.58}$&$43.88_{\pm0.54}$
&$24.54_{\pm0.48}$&$27.48_{\pm0.52}$&$29.63_{\pm0.55}$
\\

DeepEMD{~\color{gray}{[TPAMI2023]}}~\cite{ZhangCLS23} 
&ResNet-50&$36.74_{\pm0.60}$&\underline{$50.74_{\pm0.60}$}&\underline{$57.34_{\pm0.49}$}
&$29.10_{\pm0.56}$&$37.76_{\pm0.68}$&$42.55_{\pm0.70}$\\

StyleAdv{~\color{gray}{[CVPR2023]}}~\cite{FuXFJ23} 
&ResNet-50&$37.02_{\pm0.67}$&$46.36_{\pm0.57}$&$51.15_{\pm0.53}$
&\bm{$32.98_{\pm0.77}$}&\bm{$40.18_{\pm0.74}$}&$42.32_{\pm0.71}$\\

ESPT{~\color{gray}{[AAAI2023]}}~\cite{RongLSCX23} 
&ResNet-50&\underline{$37.11_{\pm0.61}$}&$49.20_{\pm0.56}$&$54.18_{\pm0.51}$
&\underline{$29.90_{\pm0.59}$}&\underline{$39.57_{\pm0.72}$}&$43.82_{\pm0.72}$\\

\textbf{CRLNet (Ours)} 
&ResNet-50&\bm{$42.14_{\pm0.68}$}&\bm{$58.96_{\pm0.59}$}&\bm{$66.30_{\pm0.56}$}
&$28.01_{\pm0.57}$&\underline{$39.54_{\pm0.69}$}&\bm{$44.14_{\pm0.71}$}\\
\hline
\hline

FewTURE{~\color{gray}{[CVPR2020]}}~\cite{YeHZS20} &ViT&$34.38_{\pm0.50}$&$44.13_{\pm0.76}$&$49.14_{\pm0.76}$
&\bm{$28.61_{\pm0.50}$}&$33.09_{\pm0.50}$&$34.00_{\pm0.50}$\\
HTCTrans{~\color{gray}{[CVPR2022]}}~\cite{HeLZZGYZ22} &ViT&\underline{$43.76_{\pm0.64}$}&$56.21_{\pm0.58}$&$61.84_{\pm0.55}$
&\bm{$28.60_{\pm0.61}$}&\underline{$37.04_{\pm0.72}$}&\underline{$40.14_{\pm0.76}$}\\
CPEA{~\color{gray}{[ICCV2023]}}~\cite{HaoH0WT023} 
&ViT&$38.47_{\pm0.65}$&\underline{$59.94_{\pm0.61}$}&\underline{$66.71_{\pm0.56}$}
&$27.10_{\pm0.54}$&$31.60_{\pm0.58}$&$32.89_{\pm0.58}$
\\

{\color{recolor}{CLIP}}{~\color{gray}{[ICML2021]}}~\cite{RadfordKHRGASAM21} &ViT-L &{$21.27_{\pm0.25}$}
&{$22.85_{\pm0.23}$}
&{$22.75_{\pm0.66}$}
&{$19.93_{\pm0.58}$}&{$ 20.03_{\pm0.53}$}&{$21.46_{\pm0.59}$}\\

{\color{recolor}{CLIP-Adapter}}{~\color{gray}{[IJCV2024]}}~\cite{GaoGZMFZLQ24} &ViT-L&{$21.73_{\pm0.75}$}
&{$29.96_{\pm0.48}$}
&{$36.70_{\pm0.75}$}
&{$20.63_{\pm0.67}$}&{$25.47_{\pm0.78}$}&{$26.47_{\pm0.40}$}\\

\textbf{CRLNet (Ours)} 
&ViT&\bm{$48.27_{\pm0.68}$}&\bm{$67.85_{\pm0.62}$}&\bm{$75.06_{\pm0.55}$}
&\underline{$28.27_{\pm0.56}$}&\bm{$43.91_{\pm0.74}$}&\bm{$50.46_{\pm0.78}$}\\
\hline
\hline
	\end{tabular}}
	\label{tab:comparison_B}
\end{table*}

\subsection{Experiment Setup}  

\noindent\textbf{Datasets.} For a fair comparison, all models were pre-trained on ILSVRC~\cite{DengDSLL009} and subsequently evaluated on the RD-FSL benchmark datasets. That is, we treated ILSVRC as the base classes and the six evaluation datasets in RD-FSL as the novel classes, ensuring no overlap in training. It's important to note that we exclude classes that are common between the pre-training and evaluation datasets.  

\noindent\textbf{Evaluated Methods.}   
{\color{mycolor}We selected both the classic and current SOTA FSL models for comparison, including methods from both inductive and transductive testing types.} To comprehensively assess their performance, we employed commonly used backbones in FSL methods, like ResNet-12, ResNet-50, and ViT. It is worth noting that we utilize publicly available codes and adhere to the default training configurations for these models. Additionally, all backbones were loaded with pre-trained parameters from ImageNet to compare against the optimal results of existing methods.  

\noindent\textbf{Testing Strategy.} During the testing phase, all methods underwent meta-testing on novel classes. The following experiments assessed accuracy using standard $N$-way-$K$-shot settings with $15$ query samples per class. Testing experiments were conducted with random sampling across $600$ episodes. The average accuracy, along with the 95\% confidence interval, was reported. In the testing of inductive methods, information leakage between query samples is not allowed, meaning that query images must be input sequentially for testing. However, transductive methods can utilize mutual learning between queries in a single batch to enhance performance. Therefore, the testing of inductive methods is more stringent and aligns better with real-world application settings.

\noindent\textbf{Implementation Details.} For a fair comparison, during each episode evaluation on each evaluation dataset, the trained parameters were loaded, and only the current support images were fine-tuned or used to guide the prediction of query images. All models were trained and tested on the same GPU. In CRLNet, the training process was parallelized on 4 $\times$ NVIDIA A800-SXM4-80GB GPUs. For all backbones, the input image size was set to $224$, with a learning rate of $0.001$ and AdamW optimizer. For every $20$ epoch, the weight decay was set to $0.05$. Specifically, when the backbone is ResNet-12, the batch size is set to $80$, while for others, it was set to $250$. Additionally, the prototype feature dimensions after the backbones were $640 \times 14 \times 14$ for ResNet-12, $2048 \times 7 \times 7$ for ResNet-50, and $384$ for ViT.


\begin{table*}[th]
 \renewcommand{\arraystretch}{1.0}
	\centering
  	\caption{The comparison experiment results between CRLNet and SOTA methods on 5-way-1-shot, 5-way-5-shot, and 5-way-10-shot settings on two datasets in the field of agriculture (Fungus and PlantVirus). The numbers in \textbf{Bold}  indicate the best performance, while the \underline{underline} one denotes the second best.}
  \scalebox{1.0}{
	\begin{tabular}{l|c|ccc|ccc}
\hline
\hline
{Method} &{Backbone}  &\multicolumn{3}{c|}{{Fungus}} &\multicolumn{3}{c}{{PlantVirus}}\\
&&1-shot &5-shot &10-shot &1-shot &5-shot &10-shot\\
\hline

MAML{~\color{gray}{[ICML2017]}}~\cite{FinnAL17} &ResNet-12&$25.65_{\pm0.54}$&$26.09_{\pm0.51}$&$26.10_{\pm0.48}$
&$46.42_{\pm0.95}$&$48.05_{\pm0.84}$&$49.54_{\pm0.84}$\\

ProtoNet{~\color{gray}{[NIPS2017]}}~\cite{SnellSZ17} &ResNet-12&$25.72_{\pm0.49}$&$38.31_{\pm0.71}$&$42.38_{\pm0.68}$
&$51.92_{\pm0.85}$&$71.48_{\pm0.87}$&$75.99_{\pm0.82}$\\

RelationNet{~\color{gray}{[CVPR2020]}}~\cite{ZhuangTYMZJX18} &ResNet-12&$27.52_{\pm0.57}$&$33.75_{\pm0.66}$&$37.00_{\pm0.61}$
&$37.64_{\pm0.87}$&$59.73_{\pm0.93}$&$62.19_{\pm0.93}$\\

IE{~\color{gray}{[CVPR2021]}}~\cite{Rizve0KS21} &ResNet-12&$28.34_{\pm0.57}$&$38.51_{\pm0.58}$&$44.51_{\pm0.61}$&$56.84_{\pm0.90}$&$78.51_{\pm0.78}$&\underline{$84.68_{\pm0.69}$}\\

DeepDBC{~\color{gray}{[CVPR2022]}}~\cite{XieLLWL22} &ResNet-12&$28.72_{\pm0.55}$&$40.63_{\pm0.76}$&$45.71_{\pm0.78}$
&$49.95_{\pm1.04}$&$72.26_{\pm0.83}$&$74.22_{\pm0.78}$\\

DeepEMD{~\color{gray}{[TPAMI2023]}}~\cite{ZhangCLS23} &ResNet-12&$31.91_{\pm0.63}$&$44.22_{\pm0.77}$&$49.84_{\pm0.78}$
&$59.19_{\pm1.04}$&\underline{$79.36_{\pm1.00}$}&$82.82_{\pm0.94}$\\

RanKDNN{~\color{gray}{[AAAI2023]}}~\cite{GuoGWFYZG23} &ResNet-12&\underline{$32.50_{\pm0.55}$}&\underline{$45.35_{\pm0.67}$}&\underline{$50.68_{\pm0.72}$}&$62.44_{\pm0.80}$&$77.42_{\pm0.70}$&$80.00_{\pm0.50}$\\

ESPT{~\color{gray}{[AAAI2023]}}~\cite{RongLSCX23} 
&ResNet-12&$32.06_{\pm0.62}$&$44.63_{\pm0.67}$&$50.14_{\pm0.68}$
&\underline{$64.72_{\pm0.97}$}&$79.14_{\pm0.82}$&$81.98_{\pm0.78}$\\

\textbf{CRLNet (Ours)} 
&ResNet-12&\bm{$35.50_{\pm0.74}$}&\bm{$51.52_{\pm0.71}$}&\bm{$58.84_{\pm0.72}$}
&\bm{$73.31_{\pm0.94}$}&\bm{$82.77_{\pm0.81}$}&\bm{$86.04_{\pm0.77}$}\\
\hline
\hline

IE{~\color{gray}{[CVPR2021]}}~\cite{Rizve0KS21} 
&ResNet-50&$30.71_{\pm0.61}$&$40.42_{\pm0.65}$&$44.36_{\pm0.65}$
&$65.92_{\pm0.96}$&\underline{$78.80_{\pm0.85}$}&\underline{$82.04_{\pm0.77}$}
\\

DeepEMD{~\color{gray}{[TPAMI2023]}}~\cite{ZhangCLS23} 
&ResNet-5&\underline{$30.98_{\pm0.60}$}&\underline{$41.98_{\pm0.67}$}&\underline{$46.51_{\pm0.71}$}
&$55.71_{\pm1.00}$&$76.09_{\pm0.86}$&$79.55_{\pm0.84}$
\\

StyleAdv{~\color{gray}{[CVPR2023]}}~\cite{FuXFJ23} 
&ResNet-5&$29.78_{\pm0.66}$&$35.96_{\pm0.64}$&$40.04_{\pm0.69}$
&\underline{$59.36_{\pm1.10}$}&$72.29_{\pm0.89}$&$72.91_{\pm0.85}$\\

ESPT{~\color{gray}{[AAAI2023]}}~\cite{RongLSCX23}  
&ResNet-5&$29.32_{\pm0.56}$&$40.62_{\pm0.68}$&$46.18_{\pm0.67}$
&$55.46_{\pm0.89}$&$73.84_{\pm0.85}$&$79.03_{\pm0.80}$\\

\textbf{CRLNet (Ours)}  
&ResNet-5&\bm{$36.90_{\pm0.72}$}&\bm{$55.23_{\pm0.71}$}&\bm{$62.90_{\pm0.70}$}
&\bm{$69.35_{\pm0.98}$}&\bm{$83.77_{\pm0.79}$}&\bm{$87.30_{\pm0.68}$}\\

\hline
\hline
FewTURE{~\color{gray}{[CVPR2020]}}~\cite{YeHZS20} &ViT&$29.29_{\pm0.50}$&$43.89_{\pm0.83}$&$47.73_{\pm0.83}$
&$53.94_{\pm0.50}$&$74.99_{\pm1.05}$&$76.96_{\pm1.02}$\\

HTCTrans{~\color{gray}{[CVPR2022]}}~\cite{HeLZZGYZ22} &ViT&$32.71_{\pm0.61}$&$34.17_{\pm0.62}$&$37.40_{\pm0.63}$
&$70.87_{\pm1.01}$&\underline{$82.92_{\pm0.87}$}&\underline{$86.03_{\pm0.82}$}\\

CPEA{~\color{gray}{[ICCV2023]}}~\cite{HaoH0WT023} 
&ViT&\underline{$33.21_{\pm0.68}$}&\underline{$48.24_{\pm0.74}$}&\underline{$54.18_{\pm0.72}$}
&\underline{$65.94_{\pm0.98}$}&$81.54_{\pm0.83}$&$83.49_{\pm0.77}$
\\
\textbf{CRLNet (Ours)} 
&ViT&\bm{$43.71_{\pm0.86}$}&\bm{$65.70_{\pm0.75}$}&\bm{$73.48_{\pm0.66}$}
&\bm{$74.69_{\pm0.96}$}&\bm{$89.01_{\pm0.66}$}&\bm{$91.62_{\pm0.55}$}\\
\hline
\hline
	\end{tabular}}
	\label{tab:comparison_C}
\end{table*}

\subsection{Comparative Experiments and Analysis}
We compared CRLNet with classical and SOTA FSL models on six datasets of the RD-FSL benchmark. For each dataset, we employed three settings: $5$-way-$1$-shot, $5$-way-$5$-shot, and $5$-way-$10$-shot. This means selecting five new classes each time, with each class having $1$, $5$, or $10$ randomly chosen support images to predict the query images. Generalization experiments were conducted on different backbones as well.  

Table~\ref{tab:comparison_A} presents the comparison results on two biology datasets, Animal and Insect. Across all backbones and settings, CRLNet outperforms the existing SOTA models. Specifically, on the Animal dataset, it surpasses the second-best by 11.68\%, 11.55\%, 12.77\%, 25.63\%, 24.68\%, 23.8\%, 28.4\%, 25.83\%, and 24.49\%. Furthermore, we also compared CRLNet with current optimal transductive methods, such as FM~\cite{WangLXHH24} and protoLP~\cite{ZhuK23}. Even though CRLNet uses an inductive testing approach where query image information does not leak between samples, its results remain optimal. This further demonstrates CRLNet's superior performance in testing on challenging images.
{\color{recolor}In the nine results on Insect, it outperforms the second-best by a range of 15.37\% to 27.24\%. Similarly, in Table~\ref{tab:comparison_B} and Table~\ref{tab:comparison_C}, except for a 3.61\% lower performance in the Mineral 1-shot setting with ResNet-12, and a 0.34\% lower performance in the OracleInscription 1-shot setting with ViT, CRLNet surpassed existing methods in the remaining $34$ results. In addition, we also tested the results of the base model CLIP~\cite{RadfordKHRGASAM21} and its fine-tuned models~\cite{GaoGZMFZLQ24}. The results show that their environmental robustness was significantly lower than that of CRLNet.}

\begin{figure*}[th]
	\centering
	\includegraphics[width=1.0\linewidth]{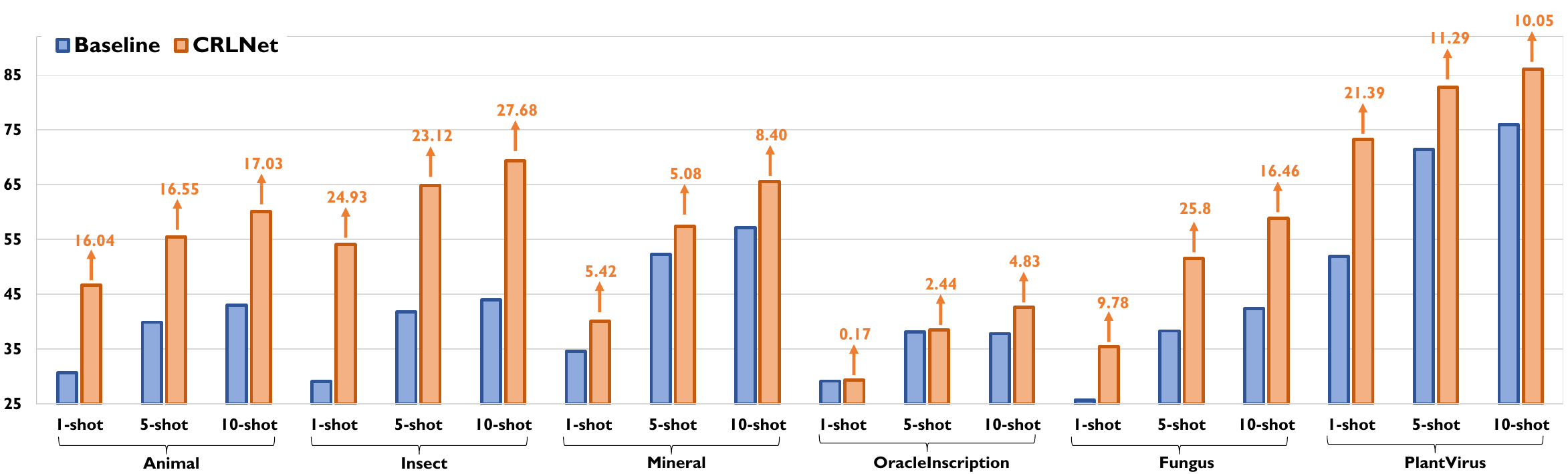}
	\caption{The comparison experiment results between CRLNet and the baseline \cite{FinnAL17} on the RD-FSL Benchmark with ResNet-12 show significant improvements. The numbers above the arrows indicate the performance enhancement of CRLNet compared to the baseline.}
	\label{Fig:result}
\end{figure*}

{\color{recolor}Figure~\ref{Fig:result} presents the performance comparison between CRLNet and its baseline method (ProtoNet~\cite{SnellSZ17}). The enhancements on the Animal dataset exceeded 16\%, with the lowest improvement on the Insect being 23.12\%, over 9\% on the Fungus, and a minimum improvement of 10.05\% on the PlantVirus. These results highlight the superiority of CRLNet, showcasing its strong generalization across multiple datasets and robust applicability to various backbones, while also demonstrating a significant enhancement in environmental robustness compared to the baseline.}

\begin{figure*}[th]
	\centering
	\includegraphics[width=1.0\linewidth]{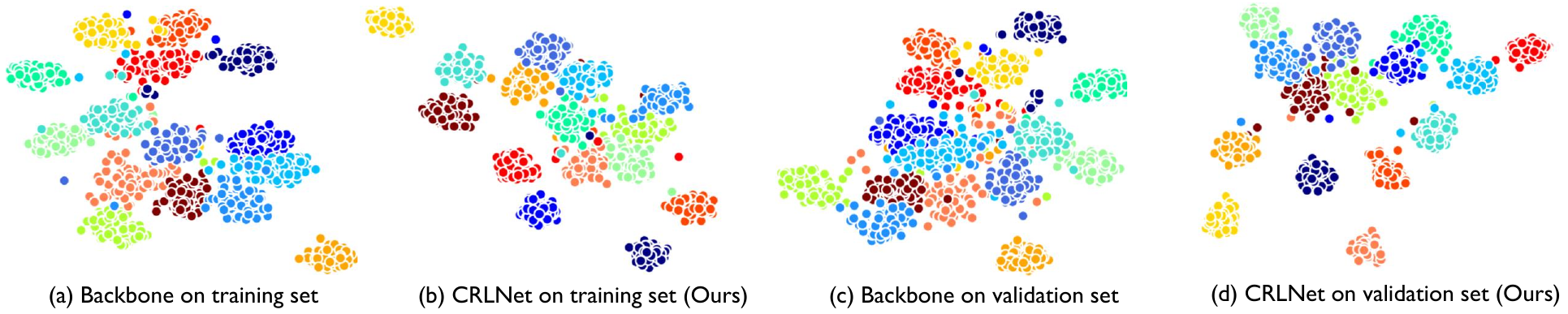}
	\caption{Visualizing the features of Backbone (ResNet-50) and CRLNet using t-SNE on both the training set (ImageNet) and the validation set (Animal). Points of the same color represent the same category. Panels (a) and (b) show that, compared to the Backbone, CRLNet learns more class-distinctive feature representations. The feature distributions in panels (c) and (d) illustrate that CRLNet demonstrates strong generalization capabilities in new domains.}
	\label{Fig:tsne}
\end{figure*}

\begin{figure*}[th]
	\centering
	\includegraphics[width=1.0\linewidth]{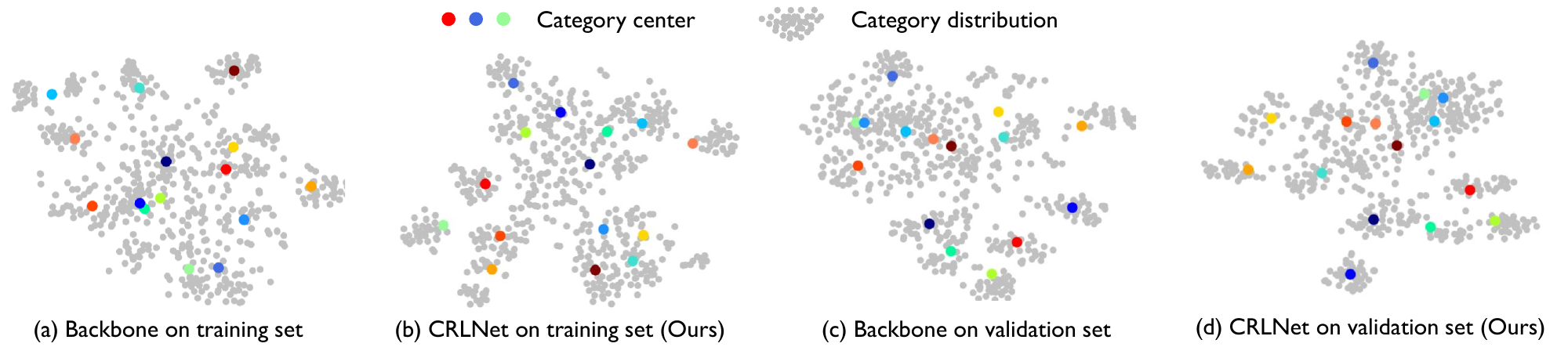}
	\caption{Comparison of category distribution and category centers obtained by Backbone and CRLNet. The comparisons between panels (a) and (b), as well as between (c) and (d), reveal that CRLNet clusters features of the same class more tightly and disperses features of different classes more effectively than the Backbone. This results in increased distances between different category centers.}
	\label{Fig:center}
\end{figure*}

\begin{figure*}[th]
	\centering
	\includegraphics[width=1.0\linewidth]{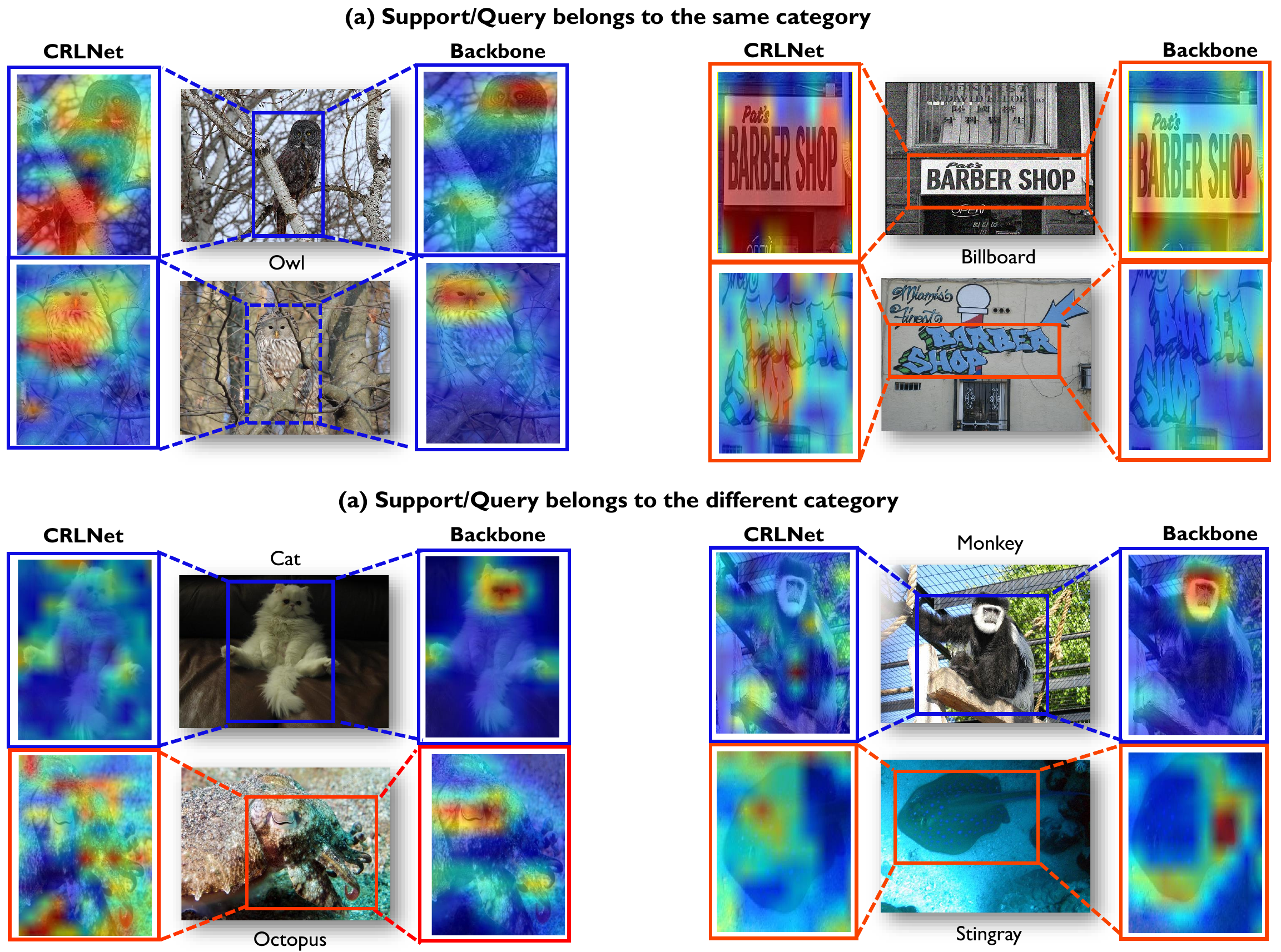}
	\caption{Feature visualization comparison between CRLNet and Backbone when support and query belong to the same category (a) or different categories (b), using the Class Activation Mapping (CAM) method~\cite{ZhouKLOT16}. Compared to the Backbone, CRLNet captures fine-grained distinguishing features within the same category and effectively disperses features of different categories.}
	\label{Fig:cam}
\end{figure*}

Based on the above results, our analysis is as follows:  
\begin{itemize}   
\item \textbf{Results on the Oraclelnscription:} Due to the majority of query images in the Oraclelnscription being incomplete font targets and blurry images, they exhibit the largest difference compared to support images. This makes it challenging for all FSL methods to quickly learn new class knowledge from limited support and to handle the extraction of query features and category prediction fused with a significant amount of noise. Even with only $1$ shot, CRLNet's performance is not satisfactory. However, as the number of provided support images increases, CRLNet's performance improvement compared to other methods is significant. This indicates that CRLNet is more adept at utilizing diverse support images to accurately identify query targets and represent their category-discriminative information.   

\item \textbf{About coarse-grained and fine-grained:} Despite using a contrastive learning framework, CRLNet focuses on utilizing important information from support images to assist query images in better representation, independent of the size of inter-class differences. Thus, CRLNet's performance is not affected.  

\item \textbf{Impact of $n$-shot:} It is commonly believed that methods based on metric learning excel in 1-shot settings, while representation learning methods thrive in n-shot settings. However, CRLNet is not influenced by this distinction. This is because, during the process of re-representation, CRLNet achieves the goal of metric learning supervision through contrastive learning loss at the feature level, aiming to minimize intra-class differences and maximize inter-class differences.  
\end{itemize}

\subsection{Visual Analysis}

We claimed that CRLNet enhances the distinguishing features of categories during the representation process of query and support images. It also utilizes contrastive learning loss to make features of the same class more tightly clustered and those of different classes more dispersed. To demonstrate CRLNet's clustering capability, we first perform t-SNE visualization of the features output by the Backbone and the features processed by CRLNet, as shown in Figure~\ref{Fig:tsne} and ~\ref{Fig:center}. 

From them, it can be observed that within the training set, even after training, the Backbone still causes overlaps between certain features, leading to errors. In contrast, CRLNet aggregates features of the same class more tightly while making the class centers of the training categories more dispersed. This effect is even more pronounced in the test set. When dealing with challenging images, {\color{recolor}the backbone tends to mix a large number of features, even when they belong to different categories. In contrast, CRLNet effectively separates these dissimilar features, maximizing the distance between class centers and minimizing overlap. These results highlight that conditional learning can leverage the generation of relational matrices to assign greater weights to important information, thereby making feature representations within the same category more similar.}  

Secondly, as shown in Figure~\ref{Fig:cam}, we visualized the Class Activation Mapping of the backbone and CRLNet. It can be observed that, compared to the Backbone, CRLNet can effectively learn more common distinguishing features from images of the same class. When dealing with images of different classes, the Backbone still persistently seeks important features in both images, but CRLNet, through feature representation guided by the relational matrix, represents features of different class images in a very dispersed manner. This visualization demonstrates that CRLNet is more adept at capturing fine distinguishing features in images of the same class and separating features of different classes.  
In summary, these visualization results confirm that CRLNet not only brings similar features closer and pushes dissimilar features further apart, but it is also adept at capturing local category features within similar images when faced with challenging samples, thereby enhancing performance.

\subsection{Ablation Experiments}

\noindent\textbf{Effectiveness of the Modules in CRLNet.} To validate the effectiveness of the conditional learner and re-representation learner, we conducted the ablation experiments as shown in Table~\ref{tab:module}. Compared to the baseline, the addition of the conditional learner resulted in significant improvements across datasets, except for the Mineral and Oracle datasets. This indicates that the direct fusion of the matrix after interaction with the conditional learner can enhance the similarity between features of the same class. However, when faced with datasets with minimal class differences, direct fusion is not as effective. The introduction of the re-representation learner compensates for the coarse representation of direct concatenation on fine-grained datasets, addressing the aforementioned issue. Nevertheless, we also observed that on some datasets, the difference between adding the re-representation learner and not adding it is minimal. This is because the coarse-grained test sets have already learned effective interaction knowledge from the conditional learner, and adding the re-representation learner does not provide a more accurate representation.
\begin{table}[ht]
	\centering
  	\caption{{\color{mycolor}The ablation study of the Conditional Learner (CL) and Re-representation Learner (RL) in CRLNet was conducted across various datasets using a 5-way-1-shot setting.}}
  \scalebox{1.0}{
	\begin{tabular}{l|c|c|c|c}
\hline
\hline
Method &Backbone&Animal &Insect &Mineral \\
\hline
Baseiline~\cite{SnellSZ17}&ResNet-12&$30.67$&$29.15$&$34.62$\\
+CL&ResNet-12&$37.98$&$44.58$&$33.18$\\
+CL+RL&ResNet-12&$46.71$&$54.08$&$40.05$\\
Baseiline~\cite{SnellSZ17}&ResNet-50&$31.80$&$32.74$&$30.11$\\
+CL&ResNet-50&$58.77$&$62.01$&$37.57$\\
+CL+RL&ResNet-50&$62.76$&$64.66$&$42.14$\\
\hline
\hline
Method&Backbone&Oracle &Fungus &Plant\\
\hline
Baseiline~\cite{SnellSZ17}&ResNet-12&$29.17$&$25.72$&$51.92$\\
+CL&ResNet-12&$27.64$&$34.53$&$63.11$\\
+CL+RL&ResNet-12&$29.34$&$35.50$&$73.31$\\
\hline
\hline
	\end{tabular}}
	\label{tab:module}
\end{table}

\begin{table}[ht]
	\centering
  	\caption{{\color{mycolor}Ablation experiments on the Different Structures in CRLNet. The results are all set in 5-way-1-shot.}}
  \scalebox{1.10}{
	\begin{tabular}{l|c|c|c|c}
\hline
\hline
Method &Animal &Insect &Mineral&Oracle\\
\hline
Non-Siamese&$48.24$&$50.16$&$37.99$&$27.50$ \\
Non-Residual&$37.19$&$26.77$&$30.22$&$22.60$\\
Residual/Siamese&$46.71$&$54.08$&$40.05$&$29.34
$\\
\hline
\hline
	\end{tabular}}
	\label{tab:structures}
\end{table}

\noindent\textbf{Ablation Experiments on Different Structures.}
We explored three different CRLNet structures and found that they had a significant impact on the results, as shown in Table~\ref{tab:structures}. The Non-Siamese structure refers to a CRLNet structure where the conditional learner and re-representation learner are independently optimized, meaning that the branches for support and query are fixed to optimize the representation networks for simple and challenging samples separately. However, the results demonstrated that the Siamese structure performed better. Analysis indicates that the Siamese structure, not constrained by fixed categories and difficulty levels of images, learned more generalizable interaction and representation capabilities. Additionally, we experimented with using only the weight matrix generated by the conditional learner for prediction, denoted as ``Non-residual''. The results showed that without the shared representation of prototype features, the weight matrix became disordered, leading to very low prediction accuracy.

\begin{table}[ht]
	\centering
  	\caption{{\color{mycolor}Ablation experiments on different data augmentation methods during image loading.}}
  \scalebox{1.10}{
	\begin{tabular}{l|c|c|c|c}
\hline
\hline
Method &Animal &Insect &Mineral &Oracle \\
\hline
None&$46.05$&$53.50$&$39.70$&$29.09$\\
+Randomcrop&$49.05$&$54.14$&$40.30$&$29.49$\\
+Noise&$46.47$&$56.19$&$40.64$&$29.27$\\
Randaugment&$46.71$&$54.08$&$40.05$&$29.34$\\
\hline
\hline
	\end{tabular}}
	\label{tab:augmentation}
\end{table}
\noindent\textbf{Ablation Experiments on Data Augmentation.}
During training, we found that applying data augmentation to query images is effective, as shown in Table~\ref{tab:augmentation}. Among them, cropping and random noise have the most significant impact on the results. This is likely because such data augmentation techniques better enable the model to learn generalization capabilities when dealing with challenging query samples.

\begin{table}[ht]
	\centering
  	\caption{{\color{mycolor}Ablation experiments on different 5-way-K-shot inference methods. }}
  \scalebox{1.0}{
	\begin{tabular}{l|c|c|c|c}
\hline
\hline
Method &Animal &Insect &Mineral &Oracle\\
\hline
Individual Similarity&$57.27$&$64.77$&$47.75$&$32.78$\\
Class Similarity&$56.10$&$65.83$&$51.96$&$32.93$\\
Classifier&$55.47$&$64.89$&$57.37$&$38.52$\\
Raw Query&$56.58$&$65.03$&$51.37$&$34.67$\\
Weighted Query&$57.09$&$67.74$&$54.06$&$35.40$\\
\hline
\hline
	\end{tabular}}
	\label{tab:shots}
\end{table}

\begin{table}[h]
	\centering
  	\caption{{\color{mycolor}Comparison experiment results of CRLNet on different foundation models.}}
  \scalebox{1.0}{
	\begin{tabular}{l|ccc|ccc}
\hline
\hline
Method &\multicolumn{3}{c|}{Animal} &\multicolumn{3}{c}{Insect}\\
&1-shot &5-shot &10-shot &1-shot &5-shot &10-shot\\
\hline
VGG-16 &$47.71$&$45.90$&$53.23$&$52.44$&$60.29$&$66.44$\\
+\textbf{CRLNet} &$48.61$&$56.51$&$59.76$&$56.27$&$66.23$&$68.92$\\
\hline
ResNet-12&$30.67$&$39.92$&$43.02$
&$29.15$&$44.28$&$47.48$\\
+\textbf{CRLNet}&$46.71$&$55.47$&$60.14$
&$54.08$&$64.89$&$69.38$ \\
\hline
ResNet-50&$31.80$&$39.52$&$44.20$
&$32.74$&$44.34$&$49.79$\\
+\textbf{CRLNet}&$62.76$&$70.79$&$72.49$
&$64.66$&$75.44$&$78.50$ \\
\hline
CLIP&$70.13$&$82.91$&$84.31$
&$71.70$&$86.12$&$88.77$\\
+\textbf{CRLNet}&$71.30$&$82.90$&$85.37$
&$73.49$&$87.24$&$89.95$ \\
\hline
Swin-T&$74.77$&$79.57$&$81.32$
&$71.47$&$79.11$&$82.36$\\
+\textbf{CRLNet}&$74.86$&$81.75$&$82.80$
&$72.37$&$81.09$&$83.41$ \\
\hline
\hline
	\end{tabular}}
	\label{tab:backbone}
\end{table}

\begin{table*}[th]
	\centering
  	\caption{{\color{mycolor}Comparison of Parameters, Computational Cost, and Performance between CRLNet and SOTA Methods (Results for 1-Shot Setting Across All Datasets).}}
  \scalebox{1.10}{
	\begin{tabular}{l|c|c|c|c|c|c|c}
\hline
\hline
Method &Backbone &Total\_P (MB)$\downarrow$ &Head\_P (MB)$\downarrow$ &Gflops$\downarrow$ &Animal  &Insect &PlantVirus\\
\hline

FEAT&RN-28 &38.12 &1.64 &82.20 &30.97&30.82&38.54\\

ESPT&ResNet-50 &23.50&1.61&16.60&37.14&49.04&55.46\\

StyleAdv &ResNet-50 &25.85 &2.34   &97.60 &32.08&33.12&59.36\\

CPEA&ViT &31.58 &21.67 &9.91 &42.46&44.54&65.94\\

EASY &ResNet-12 &37.27&24.89&7.05&30.67&42.67&62.67\\

 RanKDNN &ResNet-12 &43.04 &30.62 &182.61 &35.03 &37.99 &62.44\\
 
\textbf{CRLNet} &ResNet-12 &31.07  &18.64 &59.19 &\textbf{46.71} &\textbf{54.08} &\textbf{73.31}\\
\hline
\hline
	\end{tabular}}
	\label{tab:parameters}
\end{table*}

\noindent\textbf{$K$-Shot Inference.}
We experimented with different $K$-shot inference methods as shown in Table~\ref{tab:shots}. When predicting the query image, we first input all support images, which are $25$ support images in the $5$-way $5$-shot setting, into CRLNet to obtain $25$ similarity scores. We summed up the scores of the same class to get the result in the first row. Next, after passing through the backbone, we averaged the prototype features of the same class to get a total of $5$ class prototypes and then continue to calculate the similarity with the query image to get the result in the second row. Finally, we found that compared to the above two methods, the performance of training a classifier jointly with the 25 support features and prototype features is the best overall on all datasets, as shown in the third row.  
Furthermore, after obtaining $25$ re-represented query features interacted with support, we first used the classifier to predict the original features of the query, as shown in the fourth row. Subsequently, we averaged all obtained query features weighted with prototypes, and predicted again using the classifier, as shown in the last row. Finally, we used the features generated through support to online train the classifier for predicting the weighted features of the query as the final $K$-shot inference method of CRLNet.

{\color{mycolor}\noindent\textbf{Feasibility of CRLNet.}
Table~\ref{tab:backbone} presents the experimental results of the baseline models VGG-16, ResNet-12, ResNet-50, CLIP, and Swin-T, as well as the comparison results of CRLNet based on these models. From the experimental results, it is evident that CRLNet enhances the performance of these models across various settings. When the parameter size of the baseline models is relatively small, the improvement effect of CRLNet is particularly notable, demonstrating the accuracy of re-representation by CRLNet. When the baseline models gain strong representational capability through extensive pre-training data, such as CLIP and Swin-T, the improvements by CRLNet are smaller but can still further optimize recognition on challenging images.
}

{\color{mycolor}\noindent\textbf{Analysis of the Parameter Count and Computational Complexity of CRLNet.}
Table~\ref{tab:parameters} presents the comparison results of CRLNet with other few-shot learning methods in terms of parameter count, computational complexity, and performance. From the table, it is clear that although the newly added modules within CRLNet have slightly larger parameter sizes, when combined with smaller backbone models, CRLNet maintains a leading performance with the smallest total parameter count. Additionally, the incorporation of 4D convolution modules in CRLNet results in relatively higher computational demands when inferring support and query images separately. However, compared to other methods with even greater computational complexity, CRLNet's performance is significantly better. Nonetheless, the parameter count is an area where CRLNet requires improvement in future iterations.
}

\section{Conclusion}
\noindent\textbf{Summary.} To enhance the environmental robustness of few-shot learning methods in real-world domain scenarios, this paper introduces a new real-world multi-domain few-shot visual recognition benchmark and a novel conditional representation learning network. Firstly, the benchmark covers six datasets across four domains, where challenging and complex images encountered in real-world settings are meticulously categorized into the test dataset. Secondly, the network aims to offer a re-representation approach from a feature perspective, enabling closer intra-class and farther inter-class representations. In comparative experiments, the proposed method demonstrates superiority and strong generalization, notably surpassing baseline methods by over 10\% on multiple datasets. Ablation experiments validate the effectiveness of various modules within CRLNet and provide relevant insights.

{\color{mycolor}\noindent\textbf{Limitation.}   
Although the RD-FSL Benchmark dataset constructed in this paper already includes multiple scenarios and various real-world complex situations, there are still many extreme scenarios not covered, such as underground environments in mining and strong exposure conditions in aerospace. Therefore, in future research, we aim to incorporate more domains and more challenging real-world applications into this dataset.  
Besides, during the experiments, we found that CRLNet's use of 4D convolution requires processing five-dimensional aggregated features, which increases CRLNet's computational load. In future research, we will attempt to eliminate redundant information in the aggregated features through methods such as adaptive feature selection to further enhance the computational speed of this approach.}

\bibliographystyle{IEEEtran}
\bibliography{Ref}

\begin{thebibliography}{10}
\providecommand{\url}[1]{#1}
\csname url@samestyle\endcsname
\providecommand{\newblock}{\relax}
\providecommand{\bibinfo}[2]{#2}
\providecommand{\BIBentrySTDinterwordspacing}{\spaceskip=0pt\relax}
\providecommand{\BIBentryALTinterwordstretchfactor}{4}
\providecommand{\BIBentryALTinterwordspacing}{\spaceskip=\fontdimen2\font plus
\BIBentryALTinterwordstretchfactor\fontdimen3\font minus \fontdimen4\font\relax}
\providecommand{\BIBforeignlanguage}[2]{{%
\expandafter\ifx\csname l@#1\endcsname\relax
\typeout{** WARNING: IEEEtran.bst: No hyphenation pattern has been}%
\typeout{** loaded for the language `#1'. Using the pattern for}%
\typeout{** the default language instead.}%
\else
\language=\csname l@#1\endcsname
\fi
#2}}
\providecommand{\BIBdecl}{\relax}
\BIBdecl

\bibitem{HeZRS16}
K.~He, X.~Zhang, S.~Ren, and J.~Sun, ``Deep residual learning for image recognition,'' in \emph{2016 {IEEE} Conference on Computer Vision and Pattern Recognition, {CVPR} 2016, Las Vegas, NV, USA, June 27-30, 2016}.\hskip 1em plus 0.5em minus 0.4em\relax {IEEE} Computer Society, 2016, pp. 770--778.

\bibitem{abs-2010-11929}
A.~Dosovitskiy, L.~Beyer, A.~Kolesnikov, D.~Weissenborn, X.~Zhai, T.~Unterthiner, M.~Dehghani, M.~Minderer, G.~Heigold, S.~Gelly, J.~Uszkoreit, and N.~Houlsby, ``An image is worth 16x16 words: Transformers for image recognition at scale,'' \emph{CoRR}, vol. abs/2010.11929, 2020.

\bibitem{Zhai0HB22}
X.~Zhai, A.~Kolesnikov, N.~Houlsby, and L.~Beyer, ``Scaling vision transformers,'' in \emph{{IEEE/CVF} Conference on Computer Vision and Pattern Recognition, {CVPR} 2022, New Orleans, LA, USA, June 18-24, 2022}.\hskip 1em plus 0.5em minus 0.4em\relax {IEEE}, 2022, pp. 1204--1213.

\bibitem{Chen0CPPSGGMB0P23}
X.~Chen, X.~Wang, S.~Changpinyo, A.~J. Piergiovanni, P.~Padlewski, D.~Salz, S.~Goodman, A.~Grycner, B.~Mustafa, L.~Beyer, A.~Kolesnikov, J.~Puigcerver, N.~Ding, K.~Rong, H.~Akbari, G.~Mishra, L.~Xue, A.~V. Thapliyal, J.~Bradbury, and W.~Kuo, ``Pali: {A} jointly-scaled multilingual language-image model,'' in \emph{The Eleventh International Conference on Learning Representations, {ICLR} 2023, Kigali, Rwanda, May 1-5, 2023}.\hskip 1em plus 0.5em minus 0.4em\relax OpenReview.net, 2023.

\bibitem{VinyalsBLKW16}
O.~Vinyals, C.~Blundell, T.~Lillicrap, K.~Kavukcuoglu, and D.~Wierstra, ``Matching networks for one shot learning,'' in \emph{Advances in Neural Information Processing Systems 29: Annual Conference on Neural Information Processing Systems 2016, December 5-10, 2016, Barcelona, Spain}, D.~D. Lee, M.~Sugiyama, U.~von Luxburg, I.~Guyon, and R.~Garnett, Eds., 2016, pp. 3630--3638.

\bibitem{SnellSZ17}
J.~Snell, K.~Swersky, and R.~S. Zemel, ``Prototypical networks for few-shot learning,'' in \emph{Advances in Neural Information Processing Systems 30: Annual Conference on Neural Information Processing Systems 2017, December 4-9, 2017, Long Beach, CA, {USA}}, I.~Guyon, U.~von Luxburg, S.~Bengio, H.~M. Wallach, R.~Fergus, S.~V.~N. Vishwanathan, and R.~Garnett, Eds., 2017, pp. 4077--4087.

\bibitem{FinnAL17}
C.~Finn, P.~Abbeel, and S.~Levine, ``Model-agnostic meta-learning for fast adaptation of deep networks,'' in \emph{Proceedings of the 34th International Conference on Machine Learning, {ICML} 2017, Sydney, NSW, Australia, 6-11 August 2017}, ser. Proceedings of Machine Learning Research, D.~Precup and Y.~W. Teh, Eds., vol.~70.\hskip 1em plus 0.5em minus 0.4em\relax {PMLR}, 2017, pp. 1126--1135.

\bibitem{ChenLKWH19}
W.~Chen, Y.~Liu, Z.~Kira, Y.~F. Wang, and J.~Huang, ``A closer look at few-shot classification,'' in \emph{7th International Conference on Learning Representations, {ICLR} 2019, New Orleans, LA, USA, May 6-9, 2019}.\hskip 1em plus 0.5em minus 0.4em\relax OpenReview.net, 2019.

\bibitem{ZhuangTYMZJX18}
Y.~Zhuang, L.~Tao, F.~Yang, C.~Ma, Z.~Zhang, H.~Jia, and X.~Xie, ``Relationnet: Learning deep-aligned representation for semantic image segmentation,'' in \emph{24th International Conference on Pattern Recognition, {ICPR} 2018, Beijing, China, August 20-24, 2018}.\hskip 1em plus 0.5em minus 0.4em\relax {IEEE} Computer Society, 2018, pp. 1506--1511.

\bibitem{XieLLWL22}
J.~Xie, F.~Long, J.~Lv, Q.~Wang, and P.~Li, ``Joint distribution matters: Deep brownian distance covariance for few-shot classification,'' in \emph{{IEEE/CVF} Conference on Computer Vision and Pattern Recognition, {CVPR} 2022, New Orleans, LA, USA, June 18-24, 2022}.\hskip 1em plus 0.5em minus 0.4em\relax {IEEE}, 2022, pp. 7962--7971.

\bibitem{ZhangCLS23}
C.~Zhang, Y.~Cai, G.~Lin, and C.~Shen, ``Deepemd: Differentiable earth mover's distance for few-shot learning,'' \emph{{IEEE} Trans. Pattern Anal. Mach. Intell.}, vol.~45, no.~5, pp. 5632--5648, 2023.

\bibitem{Rizve0KS21}
M.~N. Rizve, S.~H. Khan, F.~S. Khan, and M.~Shah, ``Exploring complementary strengths of invariant and equivariant representations for few-shot learning,'' in \emph{{IEEE} Conference on Computer Vision and Pattern Recognition, {CVPR} 2021, virtual, June 19-25, 2021}.\hskip 1em plus 0.5em minus 0.4em\relax Computer Vision Foundation / {IEEE}, 2021, pp. 10\,836--10\,846.

\bibitem{GuoGWFYZG23}
Q.~Guo, H.~Gong, X.~Wei, Y.~Fu, Y.~Yu, W.~Zhang, and W.~Ge, ``Rankdnn: Learning to rank for few-shot learning,'' in \emph{Thirty-Seventh {AAAI} Conference on Artificial Intelligence, {AAAI} 2023, Thirty-Fifth Conference on Innovative Applications of Artificial Intelligence, {IAAI} 2023, Thirteenth Symposium on Educational Advances in Artificial Intelligence, {EAAI} 2023, Washington, DC, USA, February 7-14, 2023}, B.~Williams, Y.~Chen, and J.~Neville, Eds.\hskip 1em plus 0.5em minus 0.4em\relax {AAAI} Press, 2023, pp. 728--736.

\bibitem{FuXFJ23}
Y.~Fu, Y.~Xie, Y.~Fu, and Y.~Jiang, ``Styleadv: Meta style adversarial training for cross-domain few-shot learning,'' in \emph{{IEEE/CVF} Conference on Computer Vision and Pattern Recognition, {CVPR} 2023, Vancouver, BC, Canada, June 17-24, 2023}.\hskip 1em plus 0.5em minus 0.4em\relax {IEEE}, 2023, pp. 24\,575--24\,584.

\bibitem{ZhangGLZFxB24}
R.~Zhang, J.~Geng, C.~Liu, W.~Zhang, Z.~Feng, L.~Xue, and Y.~Bei, ``Multi-layer tuning {CLIP} for few-shot image classification,'' in \emph{Pattern Recognition and Computer Vision - 7th Chinese Conference, {PRCV} 2024, Urumqi, China, October 18-20, 2024, Proceedings, Part {V}}, ser. Lecture Notes in Computer Science, Z.~Lin, M.~Cheng, R.~He, K.~Ubul, W.~Silamu, H.~Zha, J.~Zhou, and C.~Liu, Eds., vol. 15035.\hskip 1em plus 0.5em minus 0.4em\relax Springer, 2024, pp. 173--186.

\bibitem{HeLZZGYZ22}
Y.~He, W.~Liang, D.~Zhao, H.~Zhou, W.~Ge, Y.~Yu, and W.~Zhang, ``Attribute surrogates learning and spectral tokens pooling in transformers for few-shot learning,'' in \emph{{IEEE/CVF} Conference on Computer Vision and Pattern Recognition, {CVPR} 2022, New Orleans, LA, USA, June 18-24, 2022}.\hskip 1em plus 0.5em minus 0.4em\relax {IEEE}, 2022, pp. 9109--9119.

\bibitem{Hu0SKH22}
S.~X. Hu, D.~Li, J.~St{\"{u}}hmer, M.~Kim, and T.~M. Hospedales, ``Pushing the limits of simple pipelines for few-shot learning: External data and fine-tuning make a difference,'' in \emph{{IEEE/CVF} Conference on Computer Vision and Pattern Recognition, {CVPR} 2022, New Orleans, LA, USA, June 18-24, 2022}.\hskip 1em plus 0.5em minus 0.4em\relax {IEEE}, 2022, pp. 9058--9067.

\bibitem{GuoCKCSSRF20}
Y.~Guo, N.~Codella, L.~Karlinsky, J.~V. Codella, J.~R. Smith, K.~Saenko, T.~Rosing, and R.~Feris, ``A broader study of cross-domain few-shot learning,'' in \emph{Computer Vision - {ECCV} 2020 - 16th European Conference, Glasgow, UK, August 23-28, 2020, Proceedings, Part {XXVII}}, ser. Lecture Notes in Computer Science, A.~Vedaldi, H.~Bischof, T.~Brox, and J.~Frahm, Eds., vol. 12372.\hskip 1em plus 0.5em minus 0.4em\relax Springer, 2020, pp. 124--141.

\bibitem{TangWH20}
L.~Tang, D.~Wertheimer, and B.~Hariharan, ``Revisiting pose-normalization for fine-grained few-shot recognition,'' in \emph{2020 {IEEE/CVF} Conference on Computer Vision and Pattern Recognition, {CVPR} 2020, Seattle, WA, USA, June 13-19, 2020}.\hskip 1em plus 0.5em minus 0.4em\relax Computer Vision Foundation / {IEEE}, 2020, pp. 14\,340--14\,349.

\bibitem{FallahMO21}
A.~Fallah, A.~Mokhtari, and A.~E. Ozdaglar, ``Generalization of model-agnostic meta-learning algorithms: Recurring and unseen tasks,'' in \emph{Advances in Neural Information Processing Systems 34: Annual Conference on Neural Information Processing Systems 2021, NeurIPS 2021, December 6-14, 2021, virtual}, M.~Ranzato, A.~Beygelzimer, Y.~N. Dauphin, P.~Liang, and J.~W. Vaughan, Eds., 2021, pp. 5469--5480.

\bibitem{QinSJ23}
X.~Qin, X.~Song, and S.~Jiang, ``Bi-level meta-learning for few-shot domain generalization,'' in \emph{{IEEE/CVF} Conference on Computer Vision and Pattern Recognition, {CVPR} 2023, Vancouver, BC, Canada, June 17-24, 2023}.\hskip 1em plus 0.5em minus 0.4em\relax {IEEE}, 2023, pp. 15\,900--15\,910.

\bibitem{ZhangLYCCC24}
L.~Zhang, Y.~Lin, X.~Yang, T.~Chen, X.~Cheng, and W.~Cheng, ``From sample poverty to rich feature learning: {A} new metric learning method for few-shot classification,'' \emph{{IEEE} Access}, vol.~12, pp. 124\,990--125\,002, 2024.

\bibitem{YangWLX22}
S.~Yang, S.~Wu, T.~Liu, and M.~Xu, ``Bridging the gap between few-shot and many-shot learning via distribution calibration,'' \emph{{IEEE} Trans. Pattern Anal. Mach. Intell.}, vol.~44, no.~12, pp. 9830--9843, 2022.

\bibitem{XuLHAS21}
J.~Xu, H.~Le, M.~Huang, S.~Athar, and D.~Samaras, ``Variational feature disentangling for fine-grained few-shot classification,'' in \emph{2021 {IEEE/CVF} International Conference on Computer Vision, {ICCV} 2021, Montreal, QC, Canada, October 10-17, 2021}.\hskip 1em plus 0.5em minus 0.4em\relax {IEEE}, 2021, pp. 8792--8801.

\bibitem{Song0CMS23}
Y.~Song, T.~Wang, P.~Cai, S.~K. Mondal, and J.~P. Sahoo, ``A comprehensive survey of few-shot learning: Evolution, applications, challenges, and opportunities,'' \emph{{ACM} Comput. Surv.}, vol.~55, no. 13s, pp. 271:1--271:40, 2023.

\bibitem{GuoDJGTWZ23}
Q.~Guo, H.~Du, X.~Jia, S.~Gao, Y.~Teng, H.~Wang, and W.~Zhang, ``Plug-and-play feature generation for few-shot medical image classification,'' in \emph{{IEEE} International Conference on Bioinformatics and Biomedicine, {BIBM} 2023, Istanbul, Turkiye, December 5-8, 2023}, X.~Jiang, H.~Wang, R.~Alhajj, X.~Hu, F.~Engel, M.~Mahmud, N.~Pisanti, X.~Cui, and H.~Song, Eds.\hskip 1em plus 0.5em minus 0.4em\relax {IEEE}, 2023, pp. 1096--1103.

\bibitem{ZhouLZWC23}
C.~Zhou, M.~Liu, S.~Zhang, P.~Wei, and B.~Chen, ``Few-shot classification of screen defects with class-agnostic mask and context-based classifier,'' \emph{{IEEE} Trans. Instrum. Meas.}, vol.~72, pp. 1--16, 2023.

\bibitem{GargS23a}
S.~Garg and P.~Singh, ``An aggregated loss function based lightweight few shot model for plant leaf disease classification,'' \emph{Multim. Tools Appl.}, vol.~82, no.~15, pp. 23\,797--23\,815, 2023.

\bibitem{BeiCZHZ24}
J.~Bei, G.~Cao, J.~Zhu, Y.~Han, and Y.~Zhang, ``Cross-domain few-shot hyperspectral image classification with bias diminishing and domain bridging,'' in \emph{{IGARSS} 2024 - 2024 {IEEE} International Geoscience and Remote Sensing Symposium, Athens, Greece, July 7-12, 2024}.\hskip 1em plus 0.5em minus 0.4em\relax {IEEE}, 2024, pp. 9961--9965.

\bibitem{ZhaTST23}
Z.~Zha, H.~Tang, Y.~Sun, and J.~Tang, ``Boosting few-shot fine-grained recognition with background suppression and foreground alignment,'' \emph{{IEEE} Trans. Circuits Syst. Video Technol.}, vol.~33, no.~8, pp. 3947--3961, 2023.

\bibitem{HuangZZXW21}
H.~Huang, J.~Zhang, J.~Zhang, J.~Xu, and Q.~Wu, ``Low-rank pairwise alignment bilinear network for few-shot fine-grained image classification,'' \emph{{IEEE} Trans. Multim.}, vol.~23, pp. 1666--1680, 2021.

\bibitem{TianX23}
P.~Tian and S.~Xie, ``An adversarial meta-training framework for cross-domain few-shot learning,'' \emph{{IEEE} Trans. Multim.}, vol.~25, pp. 6881--6891, 2023.

\bibitem{abs-2206-00092}
F.~Shakeri, M.~Boudiaf, S.~Mohammadi, I.~Sheth, M.~Havaei, I.~B. Ayed, and S.~E. Kahou, ``{FHIST:} {A} benchmark for few-shot classification of histological images,'' \emph{CoRR}, vol. abs/2206.00092, 2022.

\bibitem{PengL0LZ0H23}
Z.~Peng, M.~Luo, W.~Huang, J.~Li, Q.~Zheng, F.~Sun, and J.~Huang, ``Learning representations by graphical mutual information estimation and maximization,'' \emph{{IEEE} Trans. Pattern Anal. Mach. Intell.}, vol.~45, no.~1, pp. 722--737, 2023.

\bibitem{ChenK0H20}
T.~Chen, S.~Kornblith, M.~Norouzi, and G.~E. Hinton, ``A simple framework for contrastive learning of visual representations,'' in \emph{Proceedings of the 37th International Conference on Machine Learning, {ICML} 2020, 13-18 July 2020, Virtual Event}, ser. Proceedings of Machine Learning Research, vol. 119.\hskip 1em plus 0.5em minus 0.4em\relax {PMLR}, 2020, pp. 1597--1607.

\bibitem{HeCXLDG22}
K.~He, X.~Chen, S.~Xie, Y.~Li, P.~Doll{\'{a}}r, and R.~B. Girshick, ``Masked autoencoders are scalable vision learners,'' in \emph{{IEEE/CVF} Conference on Computer Vision and Pattern Recognition, {CVPR} 2022, New Orleans, LA, USA, June 18-24, 2022}.\hskip 1em plus 0.5em minus 0.4em\relax {IEEE}, 2022, pp. 15\,979--15\,988.

\bibitem{GidarisSK18}
S.~Gidaris, P.~Singh, and N.~Komodakis, ``Unsupervised representation learning by predicting image rotations,'' in \emph{6th International Conference on Learning Representations, {ICLR} 2018, Vancouver, BC, Canada, April 30 - May 3, 2018, Conference Track Proceedings}.\hskip 1em plus 0.5em minus 0.4em\relax OpenReview.net, 2018.

\bibitem{NorooziF16}
M.~Noroozi and P.~Favaro, ``Unsupervised learning of visual representations by solving jigsaw puzzles,'' in \emph{Computer Vision - {ECCV} 2016 - 14th European Conference, Amsterdam, The Netherlands, October 11-14, 2016, Proceedings, Part {VI}}, ser. Lecture Notes in Computer Science, B.~Leibe, J.~Matas, N.~Sebe, and M.~Welling, Eds., vol. 9910.\hskip 1em plus 0.5em minus 0.4em\relax Springer, 2016, pp. 69--84.

\bibitem{GidarisBKPC19}
S.~Gidaris, A.~Bursuc, N.~Komodakis, P.~P{\'{e}}rez, and M.~Cord, ``Boosting few-shot visual learning with self-supervision,'' in \emph{2019 {IEEE/CVF} International Conference on Computer Vision, {ICCV} 2019, Seoul, Korea (South), October 27 - November 2, 2019}.\hskip 1em plus 0.5em minus 0.4em\relax {IEEE}, 2019, pp. 8058--8067.

\bibitem{ChenGZHW21}
Z.~Chen, J.~Ge, H.~Zhan, S.~Huang, and D.~Wang, ``Pareto self-supervised training for few-shot learning,'' in \emph{{IEEE} Conference on Computer Vision and Pattern Recognition, {CVPR} 2021, virtual, June 19-25, 2021}.\hskip 1em plus 0.5em minus 0.4em\relax Computer Vision Foundation / {IEEE}, 2021, pp. 13\,663--13\,672.

\bibitem{LiuF0YLWZ21}
C.~Liu, Y.~Fu, C.~Xu, S.~Yang, J.~Li, C.~Wang, and L.~Zhang, ``Learning a few-shot embedding model with contrastive learning,'' in \emph{Thirty-Fifth {AAAI} Conference on Artificial Intelligence, {AAAI} 2021, Thirty-Third Conference on Innovative Applications of Artificial Intelligence, {IAAI} 2021, The Eleventh Symposium on Educational Advances in Artificial Intelligence, {EAAI} 2021, Virtual Event, February 2-9, 2021}.\hskip 1em plus 0.5em minus 0.4em\relax {AAAI} Press, 2021, pp. 8635--8643.

\bibitem{MaXHCGA21}
J.~Ma, H.~Xie, G.~Han, S.~Chang, A.~Galstyan, and W.~Abd{-}Almageed, ``Partner-assisted learning for few-shot image classification,'' in \emph{2021 {IEEE/CVF} International Conference on Computer Vision, {ICCV} 2021, Montreal, QC, Canada, October 10-17, 2021}.\hskip 1em plus 0.5em minus 0.4em\relax {IEEE}, 2021, pp. 10\,553--10\,562.

\bibitem{DoerschGZ20}
C.~Doersch, A.~Gupta, and A.~Zisserman, ``Crosstransformers: spatially-aware few-shot transfer,'' in \emph{Advances in Neural Information Processing Systems 33: Annual Conference on Neural Information Processing Systems 2020, NeurIPS 2020, December 6-12, 2020, virtual}, H.~Larochelle, M.~Ranzato, R.~Hadsell, M.~Balcan, and H.~Lin, Eds., 2020.

\bibitem{YangWZ22}
Z.~Yang, J.~Wang, and Y.~Zhu, ``Few-shot classification with contrastive learning,'' in \emph{Computer Vision - {ECCV} 2022 - 17th European Conference, Tel Aviv, Israel, October 23-27, 2022, Proceedings, Part {XX}}, ser. Lecture Notes in Computer Science, S.~Avidan, G.~J. Brostow, M.~Ciss{\'{e}}, G.~M. Farinella, and T.~Hassner, Eds., vol. 13680.\hskip 1em plus 0.5em minus 0.4em\relax Springer, 2022, pp. 293--309.

\bibitem{DengDSLL009}
J.~Deng, W.~Dong, R.~Socher, L.~Li, K.~Li, and L.~Fei{-}Fei, ``Imagenet: {A} large-scale hierarchical image database,'' in \emph{2009 {IEEE} Computer Society Conference on Computer Vision and Pattern Recognition {(CVPR} 2009), 20-25 June 2009, Miami, Florida, {USA}}.\hskip 1em plus 0.5em minus 0.4em\relax {IEEE} Computer Society, 2009, pp. 248--255.

\bibitem{FanJSCS020}
D.~Fan, G.~Ji, G.~Sun, M.~Cheng, J.~Shen, and L.~Shao, ``Camouflaged object detection,'' in \emph{2020 {IEEE/CVF} Conference on Computer Vision and Pattern Recognition, {CVPR} 2020, Seattle, WA, USA, June 13-19, 2020}.\hskip 1em plus 0.5em minus 0.4em\relax Computer Vision Foundation / {IEEE}, 2020, pp. 2774--2784.

\bibitem{kaggle}
``Kaggle,'' https://www.kaggle.com/.

\bibitem{AIstudio}
``Aistudio,'' https://aistudio.baidu.com/aistudio/datasetoverview.

\bibitem{AnagnostopoulouREFM23}
D.~Anagnostopoulou, G.~Retsinas, N.~Efthymiou, P.~P. Filntisis, and P.~Maragos, ``A realistic synthetic mushroom scenes dataset,'' in \emph{{IEEE/CVF} Conference on Computer Vision and Pattern Recognition, {CVPR} 2023 - Workshops, Vancouver, BC, Canada, June 17-24, 2023}.\hskip 1em plus 0.5em minus 0.4em\relax {IEEE}, 2023, pp. 6282--6289.

\bibitem{WuZLCY19}
X.~Wu, C.~Zhan, Y.~Lai, M.~Cheng, and J.~Yang, ``{IP102:} {A} large-scale benchmark dataset for insect pest recognition,'' in \emph{{IEEE} Conference on Computer Vision and Pattern Recognition, {CVPR} 2019, Long Beach, CA, USA, June 16-20, 2019}.\hskip 1em plus 0.5em minus 0.4em\relax Computer Vision Foundation / {IEEE}, 2019, pp. 8787--8796.

\bibitem{HughesS15}
D.~P. Hughes and M.~Salath{\'{e}}, ``An open access repository of images on plant health to enable the development of mobile disease diagnostics through machine learning and crowdsourcing,'' \emph{CoRR}, vol. abs/1511.08060, 2015.

\bibitem{abs-2205-09442}
M.~Wang and W.~Deng, ``Oracle-mnist: a realistic image dataset for benchmarking machine learning algorithms,'' \emph{CoRR}, vol. abs/2205.09442, 2022.

\bibitem{VelickovicCCRLB18}
P.~Velickovic, G.~Cucurull, A.~Casanova, A.~Romero, P.~Li{\`{o}}, and Y.~Bengio, ``Graph attention networks,'' in \emph{6th International Conference on Learning Representations, {ICLR} 2018, Vancouver, BC, Canada, April 30 - May 3, 2018, Conference Track Proceedings}.\hskip 1em plus 0.5em minus 0.4em\relax OpenReview.net, 2018.

\bibitem{ZhangGHS020}
S.~Zhang, S.~Guo, W.~Huang, M.~R. Scott, and L.~Wang, ``{V4D:} 4d convolutional neural networks for video-level representation learning,'' in \emph{8th International Conference on Learning Representations, {ICLR} 2020, Addis Ababa, Ethiopia, April 26-30, 2020}.\hskip 1em plus 0.5em minus 0.4em\relax OpenReview.net, 2020.

\bibitem{RongLSCX23}
Y.~Rong, X.~Lu, Z.~Sun, Y.~Chen, and S.~Xiong, ``{ESPT:} {A} self-supervised episodic spatial pretext task for improving few-shot learning,'' in \emph{Thirty-Seventh {AAAI} Conference on Artificial Intelligence, {AAAI} 2023, Thirty-Fifth Conference on Innovative Applications of Artificial Intelligence, {IAAI} 2023, Thirteenth Symposium on Educational Advances in Artificial Intelligence, {EAAI} 2023, Washington, DC, USA, February 7-14, 2023}, B.~Williams, Y.~Chen, and J.~Neville, Eds.\hskip 1em plus 0.5em minus 0.4em\relax {AAAI} Press, 2023, pp. 9596--9605.

\bibitem{LiuLPKYHY19}
Y.~Liu, J.~Lee, M.~Park, S.~Kim, E.~Yang, S.~J. Hwang, and Y.~Yang, ``Learning to propagate labels: Transductive propagation network for few-shot learning,'' in \emph{7th International Conference on Learning Representations, {ICLR} 2019, New Orleans, LA, USA, May 6-9, 2019}.\hskip 1em plus 0.5em minus 0.4em\relax OpenReview.net, 2019.

\bibitem{CuiG21}
W.~Cui and Y.~Guo, ``Parameterless transductive feature re-representation for few-shot learning,'' in \emph{Proceedings of the 38th International Conference on Machine Learning, {ICML} 2021, 18-24 July 2021, Virtual Event}, ser. Proceedings of Machine Learning Research, M.~Meila and T.~Zhang, Eds., vol. 139.\hskip 1em plus 0.5em minus 0.4em\relax {PMLR}, 2021, pp. 2212--2221.

\bibitem{BendouHLLPPG22}
Y.~Bendou, Y.~Hu, R.~Lafargue, G.~Lioi, B.~Pasdeloup, S.~Pateux, and V.~Gripon, ``Easy - ensemble augmented-shot-y-shaped learning: State-of-the-art few-shot classification with simple components,'' \emph{J. Imaging}, vol.~8, no.~7, p. 179, 2022.

\bibitem{ZhuK23}
H.~Zhu and P.~Koniusz, ``Transductive few-shot learning with prototype-based label propagation by iterative graph refinement,'' in \emph{{IEEE/CVF} Conference on Computer Vision and Pattern Recognition, {CVPR} 2023, Vancouver, BC, Canada, June 17-24, 2023}.\hskip 1em plus 0.5em minus 0.4em\relax {IEEE}, 2023, pp. 23\,996--24\,006.

\bibitem{WangLXHH24}
S.~Wang, J.~Lu, H.~Xu, Y.~Hao, and X.~He, ``Feature mixture on pre-trained model for few-shot learning,'' \emph{{IEEE} Trans. Image Process.}, vol.~33, pp. 4104--4115, 2024.

\bibitem{YeHZS20}
H.~Ye, H.~Hu, D.~Zhan, and F.~Sha, ``Few-shot learning via embedding adaptation with set-to-set functions,'' in \emph{2020 {IEEE/CVF} Conference on Computer Vision and Pattern Recognition, {CVPR} 2020, Seattle, WA, USA, June 13-19, 2020}.\hskip 1em plus 0.5em minus 0.4em\relax Computer Vision Foundation / {IEEE}, 2020, pp. 8805--8814.

\bibitem{HaoH0WT023}
F.~Hao, F.~He, L.~Liu, F.~Wu, D.~Tao, and J.~Cheng, ``Class-aware patch embedding adaptation for few-shot image classification,'' in \emph{{IEEE/CVF} International Conference on Computer Vision, {ICCV} 2023, Paris, France, October 1-6, 2023}.\hskip 1em plus 0.5em minus 0.4em\relax {IEEE}, 2023, pp. 18\,859--18\,869.

\bibitem{RadfordKHRGASAM21}
A.~Radford, J.~W. Kim, C.~Hallacy, A.~Ramesh, G.~Goh, S.~Agarwal, G.~Sastry, A.~Askell, P.~Mishkin, J.~Clark, G.~Krueger, and I.~Sutskever, ``Learning transferable visual models from natural language supervision,'' in \emph{Proceedings of the 38th International Conference on Machine Learning, {ICML} 2021, 18-24 July 2021, Virtual Event}, ser. Proceedings of Machine Learning Research, M.~Meila and T.~Zhang, Eds., vol. 139.\hskip 1em plus 0.5em minus 0.4em\relax {PMLR}, 2021, pp. 8748--8763.

\bibitem{GaoGZMFZLQ24}
P.~Gao, S.~Geng, R.~Zhang, T.~Ma, R.~Fang, Y.~Zhang, H.~Li, and Y.~Qiao, ``Clip-adapter: Better vision-language models with feature adapters,'' \emph{Int. J. Comput. Vis.}, vol. 132, no.~2, pp. 581--595, 2024.

\bibitem{ZhouKLOT16}
B.~Zhou, A.~Khosla, {\`{A}}.~Lapedriza, A.~Oliva, and A.~Torralba, ``Learning deep features for discriminative localization,'' in \emph{2016 {IEEE} Conference on Computer Vision and Pattern Recognition, {CVPR} 2016, Las Vegas, NV, USA, June 27-30, 2016}.\hskip 1em plus 0.5em minus 0.4em\relax {IEEE} Computer Society, 2016, pp. 2921--2929.

\end{thebibliography}


\begin{IEEEbiography}
[{\includegraphics[width=1in,height=1.25in,clip,keepaspectratio]{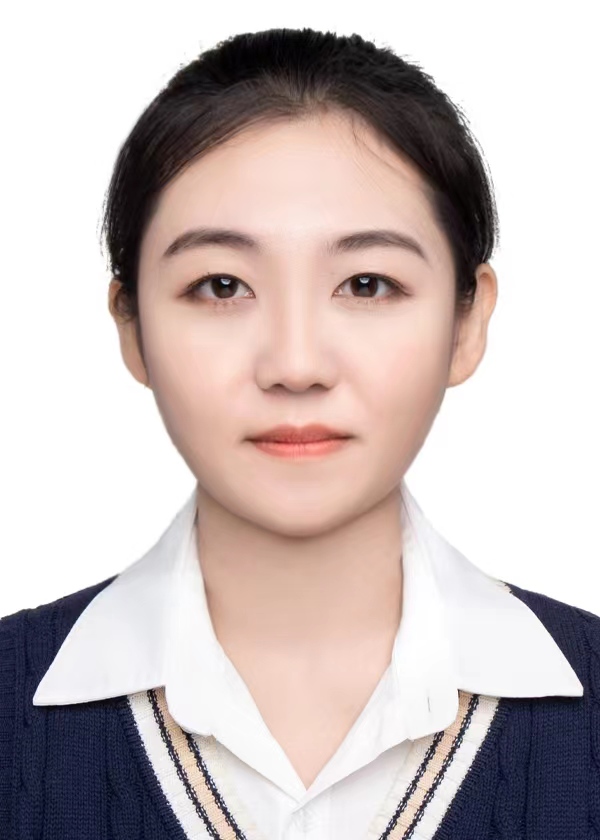}}] 
{\textbf{Qianyu Guo}} received her Ph.D. in Computer Science from Fudan University. She obtained her Master's and Bachelor's degrees from East China Normal University and Central South University, respectively. She is currently an assistant professor at Shanghai Jiao Tong University School of Medicine. Her research interests include computer vision, multi-media computing, and computer-aided diagnosis and treatment.
\end{IEEEbiography}

\begin{IEEEbiography}
[{\includegraphics[width=1in,height=1.25in,clip,keepaspectratio]{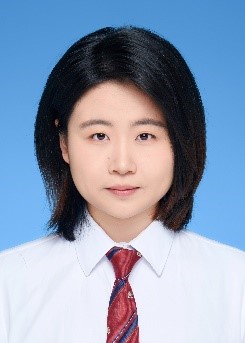}}] 
{\textbf{Jingrong Wu}} is currently pursuing a master’s degree with the School of Software Engineering, Southeast University. Her research direction is computer vision. Her research interests include semantic segmentation and image classification.
\end{IEEEbiography}

\begin{IEEEbiography}
[{\includegraphics[width=1in,height=1.25in,clip,keepaspectratio]{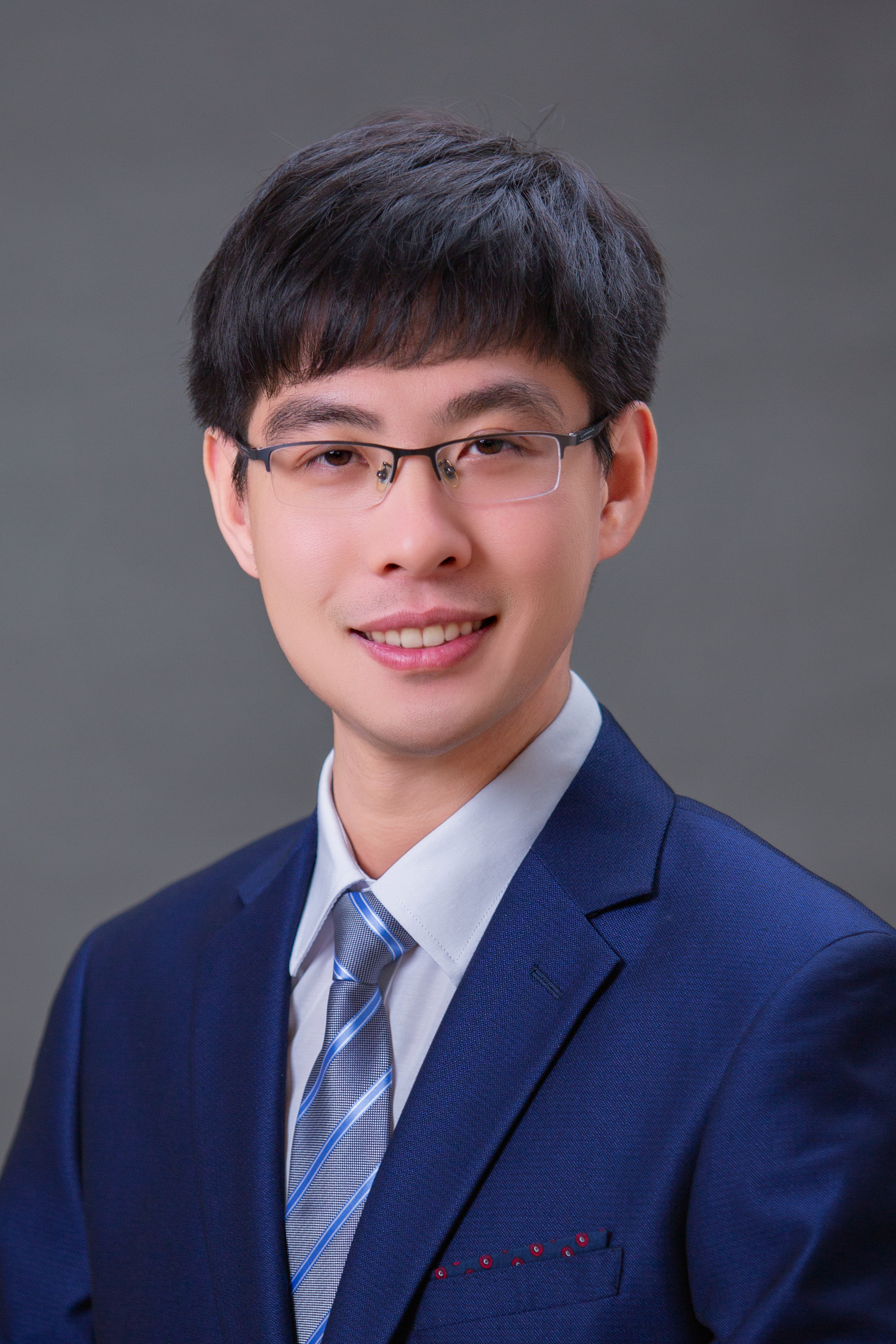}}] 
{\textbf{Tianxing Wu}} is an associate professor at the School of Computer Science and Engineering, at Southeast University. He received his Ph.D. in Software Engineering from Southeast University in 2018. His research interests include knowledge graphs, multi-modal fusion, and medical diagnosis assistance.
\end{IEEEbiography}

\begin{IEEEbiography}
[{\includegraphics[width=1in,height=1.25in,clip,keepaspectratio]{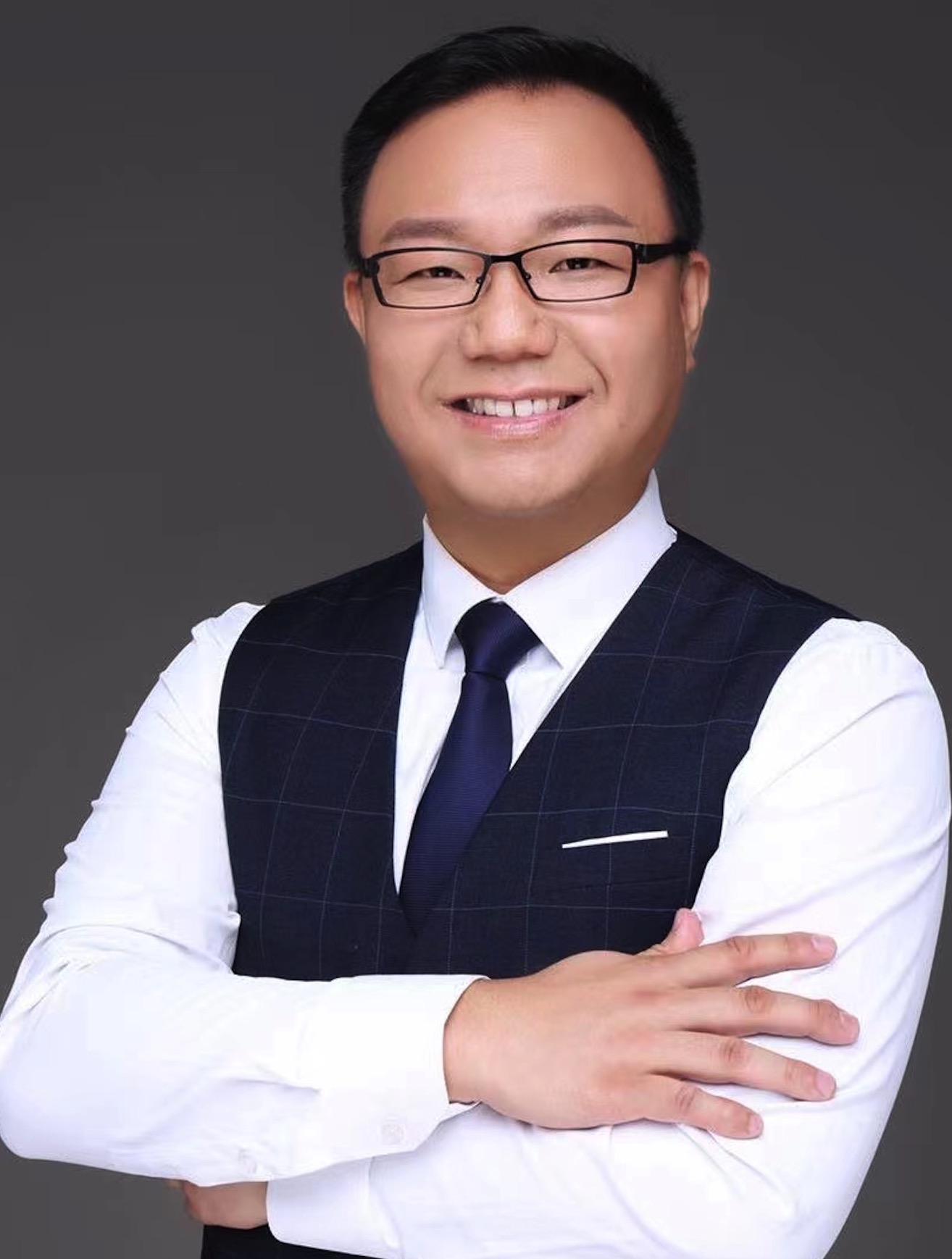}}] 
{\textbf{Haofen Wang}} is a distinguished researcher at Tongji University. He earned his Ph.D. from Shanghai Jiao Tong University in 2013. His research interests include large language models and multi-modal fusion.
\end{IEEEbiography}

\begin{IEEEbiography}
[{\includegraphics[width=1in,height=1.25in,clip,keepaspectratio]{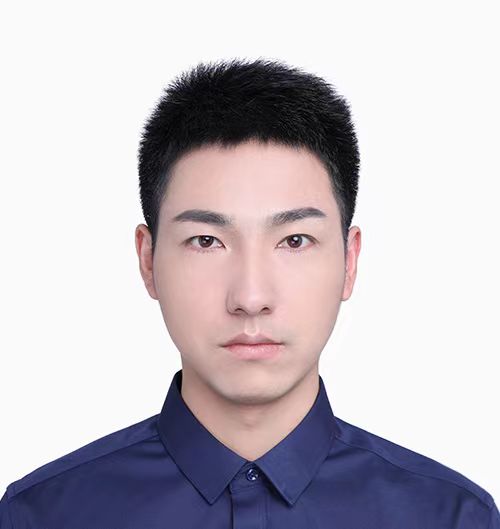}}] 
{\textbf{Weifeng Ge}} received the Ph.D. degree from The
University of Hong Kong in 2019. He is currently an Assistant Professor with the School of Computer Science, at Fudan University. His current research interests include computer vision, deep learning, artificial general intelligence, and humanoid robots.
\end{IEEEbiography}

\begin{IEEEbiography}
[{\includegraphics[width=1in,height=1.25in,clip,keepaspectratio]{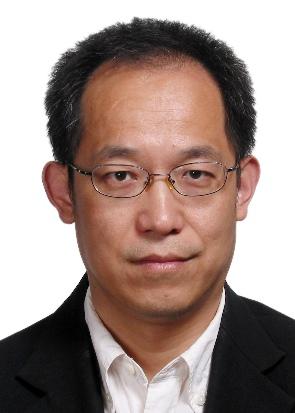}}] 
{\textbf{Wenqiang Zhang}} received a Ph.D. degree in
mechanical engineering from Shanghai Jiao Tong
University, China, in 2004. He is currently a Professor at the School of Computer Science, at Fudan University. His current research interests include computer vision and robot intelligence.
\end{IEEEbiography}

\end{document}